\documentclass{article} %
\usepackage{main,times}

\usepackage{amsmath,amsfonts,bm}

\def\eqref#1{equation~\ref{#1}}

\def\1{\bm{1}}

\DeclareMathAlphabet{\mathsfit}{\encodingdefault}{\sfdefault}{m}{sl}
\SetMathAlphabet{\mathsfit}{bold}{\encodingdefault}{\sfdefault}{bx}{n}

\usepackage{amssymb}
\usepackage{graphicx}

\usepackage{hyperref}
\usepackage{url}
\usepackage{array,booktabs,graphicx,multirow,pifont,xcolor,colortbl,capt-of}
\newcommand{\ExpTableSecondary}{\fontsize{8}{9}\selectfont}
\newcommand{\ExpTableLabel}{\fontsize{11}{12}\selectfont}
\newcommand{\ExpTableStyle}[1]{%
  \footnotesize
  \renewcommand{\arraystretch}{1.5}%
  \setlength{\extrarowheight}{0pt}%
  \setlength{\tabcolsep}{#1}%
  \setlength{\heavyrulewidth}{1.2pt}%
  \setlength{\lightrulewidth}{0.5pt}%
  \setlength{\cmidrulewidth}{0.3pt}%
  \setlength{\aboverulesep}{0.4ex}%
  \setlength{\belowrulesep}{0.65ex}%
}
\newcommand{\ExpMainTableStyle}[1]{%
  \ExpTableStyle{#1}%
  \renewcommand{\arraystretch}{1.60}%
}
\newcolumntype{N}[1]{>{\normalsize}w{c}{#1}}
\newlength{\ExpMainNumericWidth}
\newcommand{\ExpMainHeaderStrut}{\rule{0pt}{16pt}}
\newcolumntype{D}{!{\vrule width\arrayrulewidth\kern-\arrayrulewidth}}

\newsavebox{\ExpAblationFigureBox}
\newsavebox{\ExpAblationTableBox}
\newlength{\ExpAblationPanelHeight}
\providecolor{AcademicBlue}{RGB}{45,85,133}
\providecolor{EfficiencyGreen}{RGB}{46,119,78}
\providecolor{AllocationPurple}{RGB}{117,87,150}
\providecolor{CostRed}{HTML}{B85C57}
\providecolor{DynamicsBlue}{HTML}{2F5F92}
\providecolor{PolicyPurple}{HTML}{7563B4}
\providecolor{ToolTeal}{HTML}{2B7A78}

\newlength{\ExpEvolutionPanelHeight}
\usepackage{enumitem}
\usepackage{tabularx}
\usepackage[most]{tcolorbox}
\newcolumntype{Q}[1]{>{\centering\arraybackslash}m{#1}}
\providecolor{AppendixLinkBlue}{RGB}{0,85,160}
\newcommand{\AppTableFont}{\fontsize{7.5}{8}\selectfont}
\newcommand{\AppTableLabel}{\fontsize{8}{8.5}\selectfont}
\newcommand{\AppTableSecondary}{\fontsize{6.5}{7}\selectfont}
\DeclareMathSizes{7.5}{7}{5}{5}
\DeclareMathSizes{6.5}{6}{5}{5}
\newcommand{\AppTableCaptionStyle}{%
  \setlength{\abovecaptionskip}{0pt}%
  \setlength{\belowcaptionskip}{6pt}%
}
\newcommand{\AppTableStyle}{%
  \ExpMainTableStyle{3.5pt}%
  \AppTableFont
  \renewcommand{\arraystretch}{1.35}%
  \setlength{\heavyrulewidth}{0.85pt}%
  \setlength{\lightrulewidth}{0.35pt}%
  \setlength{\cmidrulewidth}{0.22pt}%
  \setlength{\arrayrulewidth}{0.28pt}%
  \setlength{\aboverulesep}{0.4ex}%
  \setlength{\belowrulesep}{0.65ex}%
  \renewcommand{\tabularxcolumn}[1]{m{##1}}%
}
\newcolumntype{Z}{>{\centering\arraybackslash}X}
\newtcolorbox{promptbox}[2][]{
  enhanced,breakable,
  colback=gray!15,colframe=black!80,
  colbacktitle=gray!30,coltitle=black,
  fontupper=\fontfamily{ptm}\selectfont\footnotesize\raggedright,
  fonttitle=\fontfamily{ptm}\selectfont\bfseries,
  title={#2},title after break={#2 (continued)},
  arc=2mm,boxrule=1.2pt,titlerule=1.2pt,
  left=10pt,right=10pt,top=6pt,bottom=6pt,toptitle=3pt,bottomtitle=3pt,#1
}
\newlist{promptitems}{itemize}{2}
\setlist[promptitems]{label=\textendash,leftmargin=1.4em,labelsep=0.45em,
  topsep=2pt,itemsep=1pt,parsep=0pt,partopsep=0pt}
\setlist[promptitems,2]{leftmargin=1.4em,topsep=1pt,itemsep=0.5pt}
\newlist{promptsteps}{enumerate}{1}
\setlist[promptsteps]{label=\arabic*.,leftmargin=1.4em,labelsep=0.4em,
  topsep=2pt,itemsep=1pt,parsep=0pt,partopsep=0pt}
\newcommand{\promptheading}[1]{\par\smallskip\noindent\textbf{#1}\par}

\title{CoEvoWhen: Policy-Tool Coevolution\\for Ultra-Long Video Temporal Grounding}

\author{Yiduo Jia$^{1}$ \enspace Muzhi Zhu$^{1}$ \enspace Jinchuan Shi$^{1}$ \enspace Hao Zhong$^{1}$ \enspace
Yuling Xi$^{1}$ \enspace Ke Liu$^{1}$ \enspace Hao Chen$^{1}$\thanks{Corresponding author.} \\[4pt]
$^{1}$Zhejiang University, State Key Lab of CAD \& CG \\[5pt]
\makebox[\dimexpr\textwidth-2\tabcolsep\relax][c]{%
  {\hypersetup{hidelinks}\color{magenta}\url{https://aim-uofa.github.io/CoEvoWhen}}}}

\hypersetup{
  pdftitle={CoEvoWhen: Policy-Tool Coevolution for Ultra-Long Video Temporal Grounding},
  pdfauthor={Yiduo Jia, Muzhi Zhu, Jinchuan Shi, Hao Zhong, Yuling Xi, Ke Liu, Hao Chen}
}

\begin{document}

\setcounter{footnote}{1}
\maketitle
\lhead{Preprint}
\vspace{-6pt}

\begin{abstract}
Ultra-long video temporal grounding requires balancing long-range evidence search with fine-grained event understanding under a limited visual budget, yet existing agentic methods still rely largely on predefined policies and tool capabilities.
Motivated by this, we propose a novel policy--tool coevolution framework that jointly evolves high-level policies and executable media tools from the agentic reasoning trajectories of a VLM, forming a reusable skill without updating model parameters.
During evolution, an external skill updater distills transferable task experience in long-video temporal grounding, accordingly refining the orchestration of long-range image-based and fine-grained video-based observations.
Alongside these policy updates, the updater employs its coding capabilities to upgrade existing tools or create new ones, adapting the tools to long-video evidence acquisition.
Equipped with the evolved skill, the VLM autonomously orchestrates tools under the guidance of the evolved policy, coordinating image and video observations for agentic inference without relying on a separate, stronger planning model.
Extensive experiments spanning five benchmarks and three VLMs show that policy--tool coevolution consistently improves temporal grounding accuracy in ultra-long videos while reducing visual token cost at inference, and that the evolved skill yields substantial performance gains on general long-video QA without additional task-specific evolution, demonstrating the effectiveness and generalizability of our framework for long-video understanding.

\end{abstract}

\setcounter{topnumber}{1}
\setcounter{dbltopnumber}{1}

\section{Introduction}
\label{sec:introduction}

Multimodal agents that support footage retrieval, clip extraction, and editing assistance must not only understand what happens in a video, but also translate users' semantic intent into concrete segments on the timeline for subsequent operations~\citep{vidi2025vidi}.
Video temporal grounding, which determines the time intervals of events described in natural language, is a fundamental capability bridging video understanding and downstream video operations~\citep{gao2017tall,lei2021qvhighlights}.
However, as videos extend to tens of minutes or even hours, target events may occupy only a tiny fraction of the timeline, so accurate grounding requires both searching for relevant content across a vast temporal range and discerning event details, action changes, and temporal boundaries locally~\citep{soldan2022mad,hannan2025revisionllm,seo2026extremewhenbench}.
Under a limited visual budget, coarse observation may overlook brief events or key details, whereas repeated fine-grained inspection incurs high visual cost~\citep{buch2025ffs,hong2025motionbench}.
Balancing long-range evidence search with fine-grained event understanding is therefore a central challenge in ultra-long video temporal grounding.

\begin{figure}[t]
    \centering
    \includegraphics[width=\linewidth]{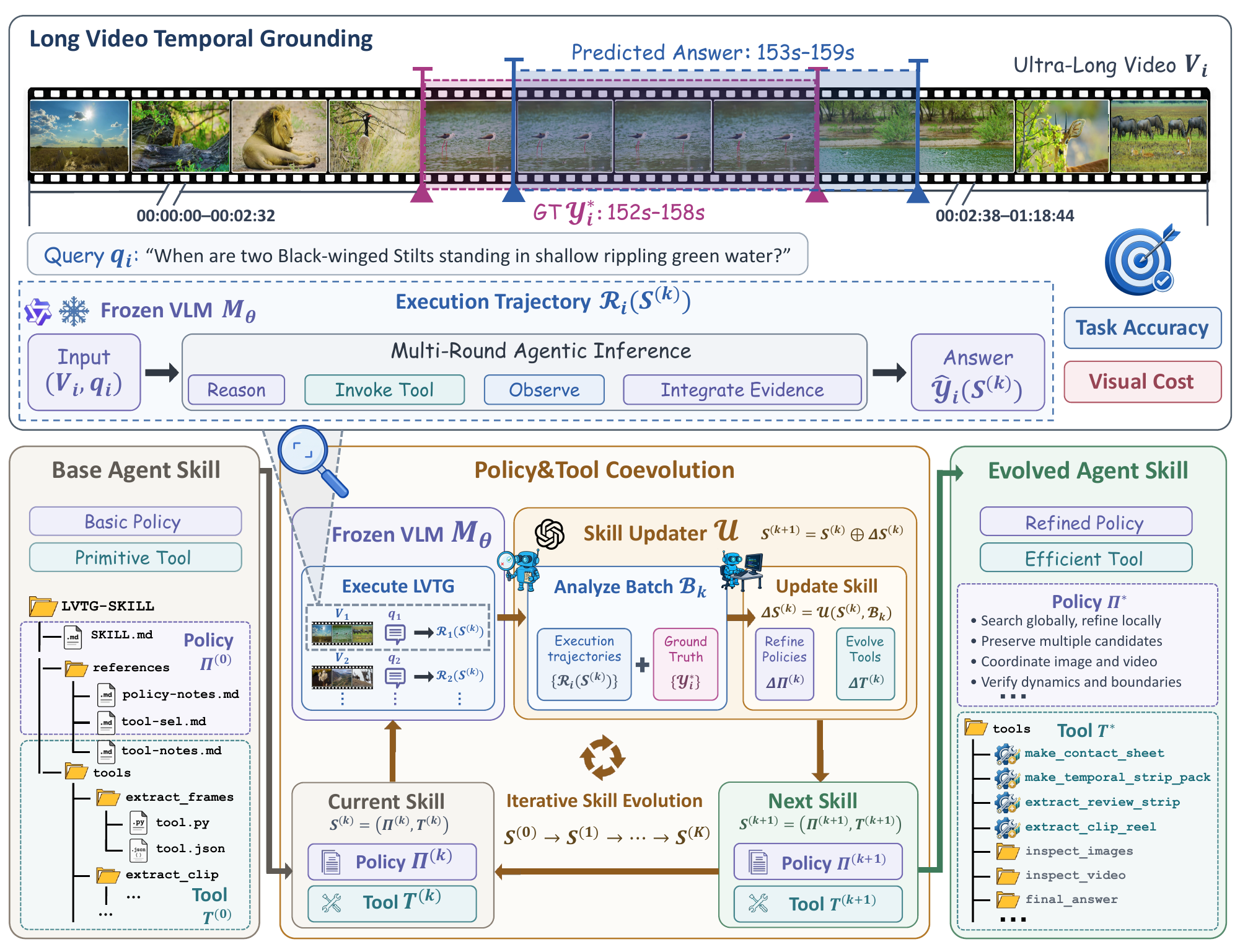}
    \caption{\textbf{Policy--tool coevolution framework for ultra-long video temporal grounding.} A frozen VLM executes grounding tasks with an external skill comprising high-level policies and executable media tools. Across evolution rounds, an external skill updater jointly refines the policies and evolves the tools from execution trajectories, synthesizing code to upgrade existing tools or create new ones.}
    \label{fig:policy_tool_coevolution}
\end{figure}

From an agentic perspective, ultra-long video temporal grounding can be organized as a process of actively acquiring, comparing, and verifying visual evidence through tool interactions, rather than a one-shot prediction on a fixed video input~\citep{liu2026videomind}.
Conditioned on the query and accumulated observations, an agent can dynamically decide which time range to search next, which candidates to examine, and whether additional evidence is needed to confirm the event and its boundaries.
Existing studies have demonstrated the potential of multi-step search and tool use, yet the underlying observation mechanisms and media processing capabilities remain largely predefined~\citep{wang2026avp}.
Tools are not merely interfaces for executing instructions; they also determine what visual input the model actually receives~\citep{hu2024sketchpad}.
Beyond studying how an agent uses existing tools, it is therefore necessary to consider how the tools themselves should be designed to better support search and verification.
Moreover, evidence needs vary across videos and queries~\citep{wu2024longvideobench}, making it difficult to predesign a general pipeline that balances grounding accuracy with visual cost.
This raises the central question of this work: \textit{how can task experience be leveraged to jointly improve a video agent's tool capabilities and the policies guiding their use, so that grounding evidence can be acquired more accurately and efficiently?}

Our key insight is that tool-use policies and tool design are interdependent: high-level policies determine what evidence is needed, when to acquire it, and how to organize the search, while media tools determine the form and granularity of the evidence the tools can present to the model.
For example, comparing temporally distant candidates calls for image-based observations that compactly cover long temporal ranges while preserving their temporal correspondence, whereas judging action order, state changes, and event continuity may require local video-based observations~\citep{ye2025temporal_search,wu2025numpro,li2024mvbench}.
Effectively exploiting these complementary capabilities requires policies that select and orchestrate observations according to evidence needs, together with tools that support the corresponding sampling and presentation.
Refining policies alone may be limited by existing tool capabilities, while expanding tools alone does not necessarily enable an agent to make effective use of the new capabilities.
We therefore explore coevolving policies and tools from the same execution feedback, enabling mutual adaptation between observation capabilities and their orchestration.
Candidate omissions, boundary errors, and dynamic ambiguities exposed in execution trajectories can inform not only adjustments to search decisions but also improvements to media tools, thereby accumulating task experience in both policies and executable capabilities.

Building upon these insights, we propose CoEvoWhen, a policy--tool coevolution framework for ultra-long video temporal grounding.
The framework jointly represents high-level policies and executable media tools as a reusable external skill, and iteratively improves it from task experience without updating the parameters of the vision-language model (VLM).
Starting from a minimal base skill that provides only the basic grounding protocol, initial observation instructions, and primitive image and video observation operations, the VLM executes grounding tasks to generate trajectories, which are paired with the corresponding ground-truth temporal intervals to form task feedback.
An external skill updater then distills transferable experience from this feedback, refining policies for task planning and observation orchestration while synthesizing code to modify existing media tools or create new ones, so that improvements extend beyond prompts and invocation sequences to the actual evidence sampling and presentation capabilities.
The evolved skill supports image-based search, candidate refinement, and selective video verification.
At inference, the VLM autonomously invokes tools from the fixed skill based on the query and accumulated evidence, without relying on a separate, stronger planning model.

We conduct systematic evaluations on three ultra-long video temporal grounding benchmarks and two long-video question answering (QA) benchmarks.
The results demonstrate that coevolution improves grounding accuracy while simultaneously reducing visual token cost.
Notably, on ExtremeWhenBench~\citep{seo2026extremewhenbench}, evolving the base skill for Qwen3.5-27B~\citep{qwen2026qwen35} raises mIoU by \textbf{74.9\%}, while reducing the average cumulative visual token cost by \textbf{11.4\%}.
Consistent gains are also observed when skills are evolved separately on different VLMs. Moreover, applying the evolved grounding skill directly to long-video QA improves performance without additional task-specific evolution, demonstrating the cross-task reusability of the accumulated experience.
Ablation studies further show that coevolution attains a better combination of accuracy and visual cost than evolving either policies or tools alone, supporting the rationale for jointly adapting observation capabilities and their usage policies.

In summary, our main contributions are as follows:
\begin{enumerate}[label=\textbf{\arabic*)}]
    \item We propose CoEvoWhen, which jointly evolves high-level policies and executable media tools from a VLM's execution trajectories, distilling ultra-long video grounding experience into a reusable external skill without updating model parameters.
    \item We evolve the orchestration of complementary image-based and video-based observations, enabling the VLM to acquire evidence autonomously without manually predefined coordination strategies or a stronger external planner.
    \item We conduct systematic experiments across five benchmarks and three VLMs, complemented by policy--tool and image--video ablations, substantiating the effectiveness and generalizability of our framework, as well as the transferability of the evolved skill to general long-video understanding.
\end{enumerate}

\section{Related Work}
\label{sec:related_work}

\subsection{Long-Video Temporal Grounding}

Video temporal grounding aims to identify temporal intervals corresponding to natural language queries in untrimmed videos~\citep{mu2024snag}.
Early methods typically rely on precomputed video features to predict these intervals in a single pass~\citep{zhang2020vslnet,moon2023qddetr}.
VTimeLLM~\citep{huang2024vtimellm}, TimeChat~\citep{ren2024timechat}, and UniTime~\citep{li2025unitime} further enhance the temporal awareness of generative multimodal models through grounding-specific adaptation, enabling direct generation of query-relevant timestamps or intervals.
As video duration grows to tens of minutes or even several hours, CONE~\citep{hou2023cone}, SOONet~\citep{pan2023soonet}, and ReVisionLLM~\citep{hannan2025revisionllm} narrow the temporal search space through query-guided window selection, single-pass scanning, and recursive refinement, respectively.
More flexible agentic frameworks acquire query-relevant visual evidence dynamically: VideoAgent~\citep{wang2024videoagent} employs an LLM agent to iteratively identify relevant visual information, VideoTree~\citep{wang2025videotree} constructs a query-adaptive hierarchical video representation, and DVD~\citep{zhang2025dvd} uses an LLM to plan and orchestrate visual tools.
Although these frameworks advance evidence acquisition from one-shot prediction to multi-step search and tool use, their underlying media processing capabilities remain largely predefined.
We instead use the VLM's temporal grounding trajectories to jointly evolve media tools and the coordination of image-based and video-based observations, enabling more accurate and efficient grounding in ultra-long videos.

\subsection{Self-Evolving Agent Skills}

Self-evolving skill frameworks turn execution trajectories and interaction feedback into reusable policies, programs, or tools, allowing task experience to accumulate without updating model parameters~\citep{shinn2023reflexion,wang2025awm,zhang2025aflow,zhang2026spatial}.
For task planning and tool use, ExpeL~\citep{zhao2024expel} distills execution experience into reusable natural language guidance, while XSkill~\citep{jiang2026xskill} organizes multimodal experience into task-level skills.
Experience can also be retained in executable form, with Voyager~\citep{wang2023voyager} accumulating successful programs as reusable code skills for retrieval and composition across tasks.
These approaches advance experience reuse, but do not explicitly couple high-level policy refinement with updates to underlying tool implementations.
Although SkillSmith~\citep{wei2026skillsmith} adapts both skills and tools, it restricts tool updates to predefined operations on the existing tool library.
For long-video understanding, META~\citep{huang2026meta} abstracts tool trajectories into reusable macro-tools and refines tool-specific usage constraints, but leaves the task-level orchestrator outside the evolution loop.
In contrast, our framework jointly evolves high-level policies and executable media tools, adapting visual sampling and presentation alongside observation orchestration to yield a reusable skill that the frozen VLM executes autonomously without relying on a stronger external planner.

\section{Method}
\label{sec:method}

\subsection{Problem Formulation}
\label{sec:problem_formulation}

Given an ultra-long video $V_i$ of duration $L_i$ and a natural language query $q_i$, video temporal grounding aims to precisely localize all temporal intervals that semantically correspond to $q_i$.
For a frozen VLM $M_\theta$ equipped with an external skill $\mathcal S$, we denote the ground-truth and predicted interval sets by
\[
\mathcal Y_i^\ast
=
\{[s_{in},e_{in}]\}_{n=1}^{N_i},
\quad
\widehat{\mathcal Y}_i(\mathcal S)
=
M_\theta(V_i,q_i;\mathcal S)
=
\{[\widehat{s}_{in},\widehat{e}_{in}]\}_{n=1}^{\widehat{N}_i},
\]
where $0\leq s_{in}<e_{in}\leq L_i$, and $N_i=0$ indicates that the target event is absent from the video. Starting from a base skill $\mathcal S^{(0)}$, we evolve it on the evolution set while keeping the model parameters $\theta$ fixed, and evaluate the resulting skill $\mathcal S^\ast$ by both grounding performance and visual token cost.

\subsection{Policy--Tool Coevolution}
\label{sec:policy_tool_coevolution}

As shown in Figure~\ref{fig:policy_tool_coevolution}, we jointly evolve the high-level policies and executable media tools within the external skill over $K$ evolution rounds.
At each evolution round, the frozen VLM uses the current skill to execute a batch of ultra-long video temporal grounding tasks.
An external skill updater then analyzes the resulting trajectories to update the policies and tools.

\subsubsection{Skill Representation}
\label{sec:skill_representation}

At evolution round $k$, the external skill $\mathcal S^{(k)}$ comprises high-level policies $\Pi^{(k)}$ and a set of executable media tools $\mathcal T^{(k)}$:
\[
\mathcal S^{(k)}
=
\left(\Pi^{(k)},\mathcal T^{(k)}\right),
\qquad
\mathcal T^{(k)}
=
\{\tau_j^{(k)}\}_{j=1}^{J_k}.
\]
$\Pi^{(k)}$ guides task planning and observation orchestration, specifying what visual evidence is needed and how it is acquired. $\mathcal T^{(k)}$ provides executable media tools adapted to diverse observation contexts.
Each tool $\tau_j=(d_j,\sigma_j,f_j)$ consists of a description $d_j$ of its capability and intended use, an interface specification $\sigma_j$, and source code $f_j$.
$J_k$ denotes the current number of tools and changes as tools are created, consolidated, or retired.

\begin{figure}[t]
    \centering
    \includegraphics[width=\linewidth]{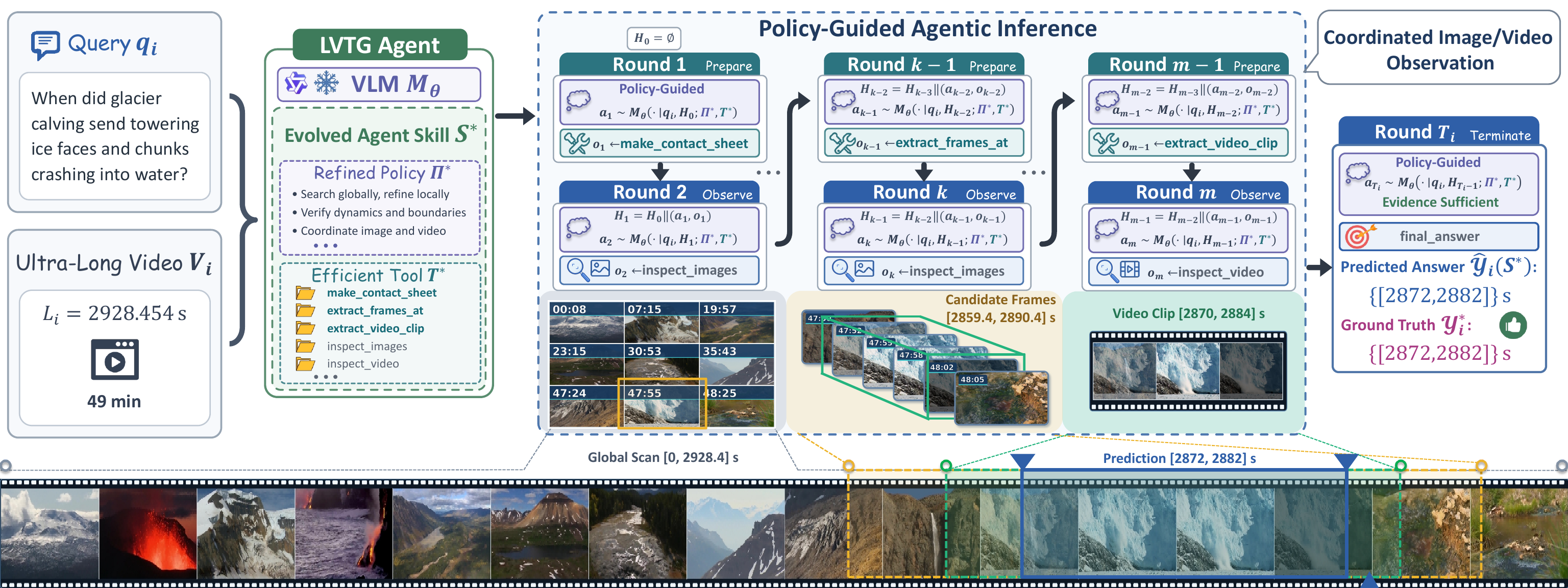}
    \caption{\textbf{Policy-guided agentic inference with the evolved skill.} Guided by the evolved policy, the VLM autonomously orchestrates media tools, using accumulated evidence to decide on subsequent observations without relying on a separate, stronger planning model. The illustrated trajectory combines image-based global search and candidate refinement with selective video verification to localize the queried event.}
    \label{fig:agentic_inference}
\end{figure}

\subsubsection{Skill Update}
\label{sec:skill_update}

Equipped with the current skill $\mathcal S^{(k)}$, the frozen VLM $M_\theta$ performs temporal grounding for a batch of queries indexed by $\mathcal I_k$.
Each trajectory $\mathcal R_i(\mathcal S^{(k)})$ records the complete execution process and final prediction for query $q_i$.
After the batch is completed, these trajectories are paired with the corresponding ground-truth intervals to form the task feedback used to update the skill:
\[
\mathcal B_k
=
\left\{\left(\mathcal R_i(\mathcal S^{(k)}),\mathcal Y_i^\ast\right)\right\}_{i\in\mathcal I_k},
\quad
\Delta\mathcal S^{(k)}
=
\mathcal U\left(\mathcal S^{(k)},\mathcal B_k\right)
=
\left(\Delta\Pi^{(k)},\Delta\mathcal T^{(k)}\right).
\]
Here, $\mathcal B_k$ denotes the feedback batch at evolution round $k$, and $\mathcal U$ denotes the external skill updater.

For policy evolution, $\Delta\Pi^{(k)}$ refines the strategies for task planning and observation orchestration with reusable experience distilled from the feedback.
For tool evolution, $\Delta\mathcal T^{(k)}$ adapts and expands the media processing capabilities available to the model by upgrading existing executable tools or creating new ones.
Existing tools evolve through updates $\Delta\tau_j^{(k)}=(\Delta d_j^{(k)},\Delta\sigma_j^{(k)},\Delta f_j^{(k)})$ to their descriptions, interfaces, and source code, while the tool set is restructured through tool creation, capability consolidation, or the retirement of unsuitable tools.
We formalize this process as:
\[
\Pi^{(k+1)}
=
\Pi^{(k)}\oplus\Delta\Pi^{(k)},
\quad
\mathcal T^{(k+1)}
=
\mathcal T^{(k)}\oplus\Delta\mathcal T^{(k)}
=
\left\{\tau_j^{(k)}\oplus\Delta\tau_j^{(k)}\right\}_{j\in\mathcal J_k^{\mathrm{ret}}}
\cup\mathcal T_k^{\mathrm{new}}.
\]
Here, $\oplus$ denotes the application of an update patch, $\mathcal J_k^{\mathrm{ret}}$ indexes the tools retained after the update, and $\mathcal T_k^{\mathrm{new}}$ denotes newly created or consolidated tools.
The skill update $\mathcal S^{(k+1)}=\mathcal S^{(k)}\oplus\Delta\mathcal S^{(k)}$ proceeds for $K$ rounds, yielding the final skill $\mathcal S^\ast=\mathcal S^{(K)}$.

\subsubsection{Coordinated Image--Video Observation}
\label{sec:image_video_observation}

We initialize coevolution with a minimal base skill $\mathcal S^{(0)}$ that provides complementary image and video observation primitives without prescribing any sophisticated strategies for task planning or observation orchestration across the two modalities. $\Pi^{(0)}$ specifies only the basic temporal grounding protocol and elementary descriptions of the available observation modes, while $\mathcal T^{(0)}$ comprises only basic image, video, and auxiliary operations.

Image-based observations can provide compact coverage of extended temporal ranges and facilitate comparisons across distant candidate regions, whereas video-based observations can preserve local temporal continuity for reasoning about motion, event order, state transitions, and temporal boundaries.
During coevolution, the tools are adapted to the observation needs exposed in execution trajectories, while the policies are refined to make effective use of the evolving capabilities.
The resulting skill coordinates image-based search and candidate refinement with selective video verification to meet varying observation needs in long-video temporal grounding.

\subsection{Policy-Guided Agentic Inference}
\label{sec:agentic_inference}

At inference, the evolved skill $\mathcal S^\ast=(\Pi^\ast,\mathcal T^\ast)$ is fixed and applied to the same VLM $M_\theta$ for ultra-long video temporal grounding.
As illustrated in Figure~\ref{fig:agentic_inference}, the VLM performs agentic inference autonomously using $\mathcal S^\ast$, without relying on a separate, stronger planning model.
For each pair $(V_i,q_i)$, $\Pi^\ast$ guides task planning and observation orchestration, while $\mathcal T^\ast$ provides the executable media tools through which the model interacts with $V_i$.

Starting from an empty interaction history $H_0=\varnothing$, the model invokes a tool from $\mathcal T^\ast$ at inference round $t$ based on the query $q_i$, the accumulated history $H_{t-1}$, and the guidance from $\Pi^\ast$.
Each round of model reasoning and tool interaction is appended to the history.
The policy-guided agentic inference process is formalized as:
\[
a_t=(j_t,\eta_t)
\sim
M_\theta
\left(
\cdot\mid q_i,H_{t-1};\Pi^\ast,\mathcal T^\ast
\right),
\quad
H_t
=
H_{t-1}\mathbin{\Vert}
\left(
a_t,
f_{j_t}^\ast(V_i;\eta_t)
\right).
\]
Here, $j_t$ indexes the selected tool in $\mathcal T^\ast$, $\eta_t$ denotes invocation arguments conforming to its interface $\sigma_{j_t}^\ast$, and $\Vert$ denotes history accumulation across inference rounds.
The maximum number of inference rounds is set to $T_{\max}$.
Once the model considers the accumulated evidence sufficient or reaches this limit, it invokes the termination tool at round $T_i\leq T_{\max}$ to submit the final prediction $\widehat{\mathcal Y}_i(\mathcal S^\ast)$.

\section{Experiments}
\label{sec:experiments}

\subsection{Experimental Setup}
\label{sec:experimental_setup}

\noindent\textbf{Benchmarks, baselines, and metrics.}
We evaluate ultra-long video temporal grounding on three benchmarks.
VUE-LVTR comprises visual queries from VUE-TR~\citep{vidi2025vidi} and VUE-TR-V2~\citep{vidi2026vidi25} restricted to videos of at least 30 minutes.
Videos in ExtremeWhenBench~\citep{seo2026extremewhenbench} average approximately 76 minutes, while the evaluation on CoMET-Bench~\citep{zou2026comet} covers multi-event grounding in videos of at least 30 minutes.
We select 100 challenging VUE-LVTR queries from distinct videos for evolution and evaluate the same evolved skill on the three benchmarks.
Results on VUE-LVTR are reported on the disjoint held-out set.
We also assess long-video QA transfer on LVBench~\citep{wang2025lvbench} and LSDBench~\citep{qu2025lsdbench}.
For baselines, we primarily compare the base and evolved skills under identical task protocols and inference settings, while also including other VLM-based and agent-based methods.
These comparisons cover the open-source VLMs Qwen3.5-27B~\citep{qwen2026qwen35}, InternVL3.5-8B~\citep{wang2025internvl35}, and TimeLens-7B~\citep{zhang2026timelens}, as well as the closed-source VLMs Gemini 2.5 Flash~\citep{gemini2025gemini25} and GPT-5.6 Luna~\citep{openai2026gpt56}, with VideoMind-7B~\citep{liu2026videomind} and EvoGround-7B~\citep{jung2026evoground} serving as agent-based baselines.
Performance on all five benchmarks is measured using their official metrics, and for efficiency, we compute cumulative visual token cost by summing visual tokens received by the VLM across rounds and averaging over queries, with separate image and video costs.
Counts of model calls, local tool calls, and visual observations characterize interaction overhead.
Detailed baseline evaluation protocols and metric definitions are provided in Appendix~\ref{app:evaluation}.

\noindent\textbf{Evolution and inference.}
We evolve the skill on Qwen3.5-27B with Codex (GPT-5.5, xhigh) as the external skill updater~\citep{openai2025codex,openai2026gpt55}.
The base skill's policy specifies only the basic grounding protocol, without predefined task planning or observation coordination, while its tool set supports only basic image and video observations, media probing, frame extraction, and clip extraction.
After each batch of four queries, Codex analyzes the VLM's execution trajectories to revise policy documents and uses its coding capabilities to modify, consolidate, or create executable media tools.
We make one pass over the evolution set, keeping VLM weights frozen throughout.
During policy-guided agentic inference, we incorporate the evolved policy into the VLM's system prompt and register the skill's tools as callable functions for multi-turn reasoning.
Both skills are evaluated with thinking enabled and greedy decoding, using at most 24 inference rounds per query.
Further implementation details can be found in Appendices~\ref{app:implementation} and~\ref{app:prompts}.

\providecolor{AcademicBlue}{RGB}{45,85,133}
\begin{table*}[!t]
  \centering
  \setlength{\abovecaptionskip}{0pt}
  \setlength{\belowcaptionskip}{6pt}
  \caption{\textbf{Gains from policy--tool coevolution on ultra-long video temporal grounding.} \textbf{Bold}: best; \textcolor{AcademicBlue}{blue}: gains over the base skill, with \textcolor{AcademicBlue}{$\uparrow$} indicating relative improvements.}
  \label{tab:grounding-main}
  \ExpMainTableStyle{3.5pt}
  \resizebox{\textwidth}{!}{%
  \begin{tabular}{@{}l>{\centering\arraybackslash}m{1.8cm}|*{4}{N{\ExpMainNumericWidth}}D*{2}{N{\ExpMainNumericWidth}}D*{4}{N{\ExpMainNumericWidth}}@{}}
    \toprule[1.2pt]
      & & \multicolumn{4}{cD}{\textbf{VUE-LVTR}}
      & \multicolumn{2}{cD}{\textbf{ExtremeWhenBench}}
      & \multicolumn{4}{c}{\textbf{CoMET-Bench}} \\
      & & \multicolumn{4}{cD}{\shortstack{\ExpTableSecondary 30.1--105.5 min\\[-1pt]\ExpTableSecondary (mean 50.9 min)}}
      & \multicolumn{2}{cD}{\shortstack{\ExpTableSecondary 45.0--542.5 min\\[-1pt]\ExpTableSecondary (mean 75.8 min)}}
      & \multicolumn{4}{c}{\shortstack{\ExpTableSecondary 30.0--123.7 min\\[-1pt]\ExpTableSecondary (mean 50.5 min)}} \\
    \cmidrule(lr){3-6}\cmidrule(lr){7-8}\cmidrule(l){9-12}
      \ExpMainHeaderStrut{\ExpTableLabel\textbf{Method}}
      & {\ExpTableLabel\textbf{Setting}}
      & \footnotesize\shortstack{\textbf{\ExpTableSecondary Precision}\\[-1pt]\textbf{\ExpTableSecondary AUC} $\uparrow$}
      & \footnotesize\shortstack{\textbf{\ExpTableSecondary Recall}\\[-1pt]\textbf{\ExpTableSecondary AUC} $\uparrow$}
      & \footnotesize\shortstack{\textbf{\ExpTableSecondary IoU}\\[-1pt]\textbf{\ExpTableSecondary AUC} $\uparrow$}
      & \footnotesize\shortstack{\textbf{\ExpTableSecondary IoU}\\[-1pt]\textbf{\ExpTableSecondary @0.5} $\uparrow$}
      & \footnotesize\shortstack{\textbf{\ExpTableSecondary Mean}\\[-1pt]\textbf{\ExpTableSecondary IoU} $\uparrow$}
      & \footnotesize\shortstack{\textbf{\ExpTableSecondary Recall}\\[-1pt]\textbf{\ExpTableSecondary @0.5} $\uparrow$}
      & \footnotesize\shortstack{\textbf{\ExpTableSecondary Mean}\\[-1pt]\textbf{\ExpTableSecondary IoU} $\uparrow$}
      & \footnotesize\shortstack{\textbf{\ExpTableSecondary Recall}\\[-1pt]\textbf{\ExpTableSecondary @0.5} $\uparrow$}
      & \footnotesize\shortstack{\textbf{\ExpTableSecondary F1}\\[-1pt]\textbf{\ExpTableSecondary @0.5} $\uparrow$}
      & \footnotesize\shortstack{\textbf{\ExpTableSecondary Rejection}\\[-1pt]\textbf{\ExpTableSecondary F1} $\uparrow$} \\
    \midrule
    Qwen3.5-27B             & {\ExpTableSecondary 768 frames}          & 0.2645 & 0.2657 & 0.1904 & 0.1889 & 0.0392 & 0.0233 & 0.0683 & 0.0708 & 0.0455 & 55.05 \\
    InternVL3.5-8B          & {\ExpTableSecondary 128 frames}          & 0.1452 & 0.1649 & 0.0804 & 0.0619 & 0.0048 & 0.0018 & 0.0098 & 0.0037 & 0.0018 & 57.18 \\
    TimeLens-7B             & {\ExpTableSecondary 384 frames}          & 0.3665 & 0.3873 & 0.2758 & 0.2866 & 0.1024 & 0.0612 & 0.0277 & 0.0129 & 0.0126 & 72.17 \\
    Gemini 2.5 Flash        & {\ExpTableSecondary 128 frames}          & 0.1670 & 0.1815 & 0.0907 & 0.0684 & 0.0089 & 0.0057 & 0.0498 & 0.0369 & 0.0088 & 57.41 \\
    GPT-5.6 Luna            & {\ExpTableSecondary 128 frames}          & 0.3114 & 0.4347 & 0.2510 & 0.2280 & 0.0444 & 0.0202 & 0.0573 & 0.0310 & 0.0309 & 41.46 \\
    \midrule
    VideoMind-7B            & {\ExpTableSecondary Agentic}             & 0.2968 & 0.4223 & 0.2000 & 0.1629 & 0.0221 & 0.0000 & 0.0185 & 0.0059 & 0.0074 & 0.00 \\
    EvoGround-7B
                              & {\ExpTableSecondary Agentic}             & 0.0659 & 0.1174 & 0.0495 & 0.0423 & 0.0050 & 0.0075 & 0.0170 & 0.0113 & 0.0113 & 43.65 \\
    \midrule
    \rowcolor{gray!15}\multicolumn{2}{@{}l|}{Qwen3.5-27B + Base Skill}
      & 0.4706 & 0.4789 & 0.4107 & 0.4267 & 0.1507 & 0.1478 & 0.1355 & 0.1440 & 0.1529 & 66.13 \\
    \multicolumn{2}{@{}l|}{\textbf{Qwen3.5-27B + Evolved Skill}}
      & \textbf{0.5918} & \textbf{0.5773} & \textbf{0.5137} & \textbf{0.5603}
      & \textbf{0.2636} & \textbf{0.2785}
      & \textbf{0.1635} & \textbf{0.1702} & \textbf{0.1831} & \textbf{77.06} \\
    \multicolumn{2}{@{}l|}{\textbf{Evolution Gain}}
      & {\color{AcademicBlue}\textbf{+0.1212}} & {\color{AcademicBlue}\textbf{+0.0984}}
      & {\color{AcademicBlue}\textbf{+0.1030}} & {\color{AcademicBlue}\textbf{+0.1336}}
      & {\color{AcademicBlue}\textbf{+0.1129}} & {\color{AcademicBlue}\textbf{+0.1307}}
      & {\color{AcademicBlue}\textbf{+0.0280}} & {\color{AcademicBlue}\textbf{+0.0262}}
      & {\color{AcademicBlue}\textbf{+0.0302}} & {\color{AcademicBlue}\textbf{+10.93}} \\[-5pt]
      \multicolumn{2}{@{}l|}{}
      & {\color{AcademicBlue}\textbf{\ExpTableSecondary$\uparrow$25.8\%}} & {\color{AcademicBlue}\textbf{\ExpTableSecondary$\uparrow$20.5\%}}
      & {\color{AcademicBlue}\textbf{\ExpTableSecondary$\uparrow$25.1\%}} & {\color{AcademicBlue}\textbf{\ExpTableSecondary$\uparrow$31.3\%}}
      & {\color{AcademicBlue}\textbf{\ExpTableSecondary$\uparrow$74.9\%}} & {\color{AcademicBlue}\textbf{\ExpTableSecondary$\uparrow$88.4\%}}
      & {\color{AcademicBlue}\textbf{\ExpTableSecondary$\uparrow$20.7\%}} & {\color{AcademicBlue}\textbf{\ExpTableSecondary$\uparrow$18.2\%}}
      & {\color{AcademicBlue}\textbf{\ExpTableSecondary$\uparrow$19.8\%}} & {\color{AcademicBlue}\textbf{\ExpTableSecondary$\uparrow$16.5\%}} \\
    \bottomrule[1.2pt]
  \end{tabular}%
  }
\end{table*}

\providecolor{EfficiencyGreen}{RGB}{46,119,78}
\providecolor{AllocationPurple}{RGB}{117,87,150}
\begin{table*}[!t]
  \centering
  \setlength{\abovecaptionskip}{0pt}
  \setlength{\belowcaptionskip}{6pt}
  \caption{\textbf{Inference efficiency with the base and evolved skills.} \textcolor{EfficiencyGreen}{Green ($\downarrow$)} and \textcolor{AllocationPurple}{purple ($\uparrow$)} indicate decreases and increases relative to the base skill.}
  \label{tab:efficiency}
  \ExpMainTableStyle{3.5pt}
  \resizebox{\textwidth}{!}{%
  \begin{tabular}{@{}>{\centering\arraybackslash}m{3.45cm}|l|*{3}{N{1.29cm}}|*{5}{N{1.29cm}}@{}}
    \toprule[1.2pt]
      & & \multicolumn{3}{c|}{\shortstack{\textbf{Token Efficiency}\\[-1pt]\ExpTableSecondary (k/query)}}
      & \multicolumn{5}{c}{\shortstack{\textbf{Interaction Efficiency}\\[-1pt]\ExpTableSecondary (count/query)}} \\
    \cmidrule(lr){3-5}\cmidrule(l){6-10}
    \noalign{\vbox to 0pt{\vss\hbox{%
      \hskip\dimexpr3.45cm+\tabcolsep\relax
      \vrule width\arrayrulewidth
        height\dimexpr\aboverulesep+\cmidrulewidth+\belowrulesep\relax depth0pt}}}
      {\ExpTableLabel\textbf{Benchmark}} & {\ExpTableLabel\textbf{Method}}
      & \footnotesize\shortstack{\textbf{\ExpTableSecondary Visual}\\[-1pt]\textbf{\ExpTableSecondary Tokens} $\downarrow$}
      & \footnotesize\shortstack{\textbf{\ExpTableSecondary Image}\\[-1pt]\textbf{\ExpTableSecondary Tokens}}
      & \footnotesize\shortstack{\textbf{\ExpTableSecondary Video}\\[-1pt]\textbf{\ExpTableSecondary Tokens}}
      & \footnotesize\shortstack{\textbf{\ExpTableSecondary Model}\\[-1pt]\textbf{\ExpTableSecondary Calls}}
      & \footnotesize\shortstack{\textbf{\ExpTableSecondary Local Tool}\\[-1pt]\textbf{\ExpTableSecondary Calls}}
      & \footnotesize\shortstack{\textbf{\ExpTableSecondary Visual}\\[-1pt]\textbf{\ExpTableSecondary Obs.}}
      & \footnotesize\shortstack{\textbf{\ExpTableSecondary Image}\\[-1pt]\textbf{\ExpTableSecondary Obs.}}
      & \footnotesize\shortstack{\textbf{\ExpTableSecondary Video}\\[-1pt]\textbf{\ExpTableSecondary Obs.}} \\
    \midrule

    \cellcolor{white}\multirow{4}{*}{\shortstack{{\ExpTableLabel VUE-LVTR}\\[2pt]\ExpTableSecondary 30.1--105.5 min\\[-1pt]\ExpTableSecondary (mean 50.9 min)}}
      & \cellcolor{gray!15}Qwen3.5-27B + Base Skill
      & \cellcolor{gray!15}202.58 & \cellcolor{gray!15}162.01 & \cellcolor{gray!15}40.57
      & \cellcolor{gray!15}17.03 & \cellcolor{gray!15}8.16 & \cellcolor{gray!15}8.11
      & \cellcolor{gray!15}7.35 & \cellcolor{gray!15}0.76 \\
      & \textbf{Qwen3.5-27B + Evolved Skill}
      & \textbf{141.89} & \textbf{141.37} & \textbf{0.52}
      & \textbf{9.63} & \textbf{4.79} & \textbf{3.86} & \textbf{3.82} & \textbf{0.04} \\
      & \textbf{$\Delta$ (Evolved $-$ Base)}
      & {\color{EfficiencyGreen}\textbf{$-$60.69}}
      & {\color{EfficiencyGreen}\textbf{$-$20.64}} & {\color{EfficiencyGreen}\textbf{$-$40.05}}
      & {\color{EfficiencyGreen}\textbf{$-$7.40}} & {\color{EfficiencyGreen}\textbf{$-$3.37}}
      & {\color{EfficiencyGreen}\textbf{$-$4.25}} & {\color{EfficiencyGreen}\textbf{$-$3.53}}
      & {\color{EfficiencyGreen}\textbf{$-$0.72}} \\[-5pt]
      & 
      & {\color{EfficiencyGreen}\textbf{\ExpTableSecondary$\downarrow$29.96\%}}
      & {\color{EfficiencyGreen}\textbf{\ExpTableSecondary$\downarrow$12.74\%}}
      & {\color{EfficiencyGreen}\textbf{\ExpTableSecondary$\downarrow$98.72\%}}
      & {\color{EfficiencyGreen}\textbf{\ExpTableSecondary$\downarrow$43.5\%}}
      & {\color{EfficiencyGreen}\textbf{\ExpTableSecondary$\downarrow$41.3\%}}
      & {\color{EfficiencyGreen}\textbf{\ExpTableSecondary$\downarrow$52.4\%}}
      & {\color{EfficiencyGreen}\textbf{\ExpTableSecondary$\downarrow$48.0\%}}
      & {\color{EfficiencyGreen}\textbf{\ExpTableSecondary$\downarrow$94.7\%}}
      \\
    \midrule

    \cellcolor{white}\multirow{4}{*}{\shortstack{{\ExpTableLabel ExtremeWhenBench}\\[2pt]\ExpTableSecondary 45.0--542.5 min\\[-1pt]\ExpTableSecondary (mean 75.8 min)}}
      & \cellcolor{gray!15}Qwen3.5-27B + Base Skill
      & \cellcolor{gray!15}227.10 & \cellcolor{gray!15}179.26 & \cellcolor{gray!15}47.84
      & \cellcolor{gray!15}16.65 & \cellcolor{gray!15}8.98 & \cellcolor{gray!15}6.93
      & \cellcolor{gray!15}5.69 & \cellcolor{gray!15}1.24 \\
      & \textbf{Qwen3.5-27B + Evolved Skill}
      & \textbf{201.17} & 197.49 & \textbf{3.68}
      & \textbf{11.19} & \textbf{5.69} & \textbf{4.56} & \textbf{4.35} & \textbf{0.21} \\
      & \textbf{$\Delta$ (Evolved $-$ Base)}
      & {\color{EfficiencyGreen}\textbf{$-$25.93}}
      & {\color{AllocationPurple}+18.23} & {\color{EfficiencyGreen}\textbf{$-$44.16}}
      & {\color{EfficiencyGreen}\textbf{$-$5.46}} & {\color{EfficiencyGreen}\textbf{$-$3.29}}
      & {\color{EfficiencyGreen}\textbf{$-$2.37}} & {\color{EfficiencyGreen}\textbf{$-$1.34}}
      & {\color{EfficiencyGreen}\textbf{$-$1.03}} \\[-5pt]
      &
      & {\color{EfficiencyGreen}\textbf{\ExpTableSecondary$\downarrow$11.42\%}}
      & {\color{AllocationPurple}\ExpTableSecondary$\uparrow$10.17\%}
      & {\color{EfficiencyGreen}\textbf{\ExpTableSecondary$\downarrow$92.31\%}}
      & {\color{EfficiencyGreen}\textbf{\ExpTableSecondary$\downarrow$32.8\%}}
      & {\color{EfficiencyGreen}\textbf{\ExpTableSecondary$\downarrow$36.6\%}}
      & {\color{EfficiencyGreen}\textbf{\ExpTableSecondary$\downarrow$34.2\%}}
      & {\color{EfficiencyGreen}\textbf{\ExpTableSecondary$\downarrow$23.6\%}}
      & {\color{EfficiencyGreen}\textbf{\ExpTableSecondary$\downarrow$83.1\%}}
      \\
    \midrule

    \cellcolor{white}\multirow{4}{*}{\shortstack{{\ExpTableLabel CoMET-Bench}\\[2pt]\ExpTableSecondary 30.0--123.7 min\\[-1pt]\ExpTableSecondary (mean 50.5 min)}}
      & \cellcolor{gray!15}Qwen3.5-27B + Base Skill
      & \cellcolor{gray!15}119.66 & \cellcolor{gray!15}67.57 & \cellcolor{gray!15}52.09
      & \cellcolor{gray!15}11.71 & \cellcolor{gray!15}6.51 & \cellcolor{gray!15}4.63
      & \cellcolor{gray!15}3.01 & \cellcolor{gray!15}1.62 \\
      & \textbf{Qwen3.5-27B + Evolved Skill}
      & \textbf{97.10} & 91.65 & \textbf{5.45}
      & \textbf{8.88} & \textbf{4.77} & \textbf{3.42} & 3.07 & \textbf{0.35} \\
      & \textbf{$\Delta$ (Evolved $-$ Base)}
      & {\color{EfficiencyGreen}\textbf{$-$22.56}}
      & {\color{AllocationPurple}+24.08} & {\color{EfficiencyGreen}\textbf{$-$46.64}}
      & {\color{EfficiencyGreen}\textbf{$-$2.83}} & {\color{EfficiencyGreen}\textbf{$-$1.74}}
      & {\color{EfficiencyGreen}\textbf{$-$1.21}} & {\color{AllocationPurple}+0.06}
      & {\color{EfficiencyGreen}\textbf{$-$1.27}} \\[-5pt]
      &
      & {\color{EfficiencyGreen}\textbf{\ExpTableSecondary$\downarrow$18.85\%}}
      & {\color{AllocationPurple}\ExpTableSecondary$\uparrow$35.64\%}
      & {\color{EfficiencyGreen}\textbf{\ExpTableSecondary$\downarrow$89.54\%}}
      & {\color{EfficiencyGreen}\textbf{\ExpTableSecondary$\downarrow$24.2\%}}
      & {\color{EfficiencyGreen}\textbf{\ExpTableSecondary$\downarrow$26.7\%}}
      & {\color{EfficiencyGreen}\textbf{\ExpTableSecondary$\downarrow$26.1\%}}
      & {\color{AllocationPurple}\ExpTableSecondary$\uparrow$2.0\%}
      & {\color{EfficiencyGreen}\textbf{\ExpTableSecondary$\downarrow$78.4\%}}
      \\
    \bottomrule[1.2pt]
  \end{tabular}%
  }
\end{table*}

\subsection{Main Results}
\label{sec:main_results}

\subsubsection{Ultra-Long Video Temporal Grounding}
\label{sec:grounding_results}

As shown in Table~\ref{tab:grounding-main}, the evolved skill achieves the best results on all reported metrics, surpassing seven VLM-based and agent-based baselines.
Relative to the base skill, IoU AUC rises from 0.4107 to 0.5137 on the VUE-LVTR held-out set.
For ExtremeWhenBench, evolution further improves mIoU and Recall@0.5 by \textbf{74.9\%} and \textbf{88.4\%}, respectively, suggesting that the evolved skill supports effective evidence search in hour-long videos and generalizes beyond the evolution data.
Results on CoMET-Bench show gains of \textbf{0.0280} in mIoU and \textbf{10.93} points in Rejection-F1, indicating that the coevolved observation capabilities help the frozen VLM acquire, verify, and select evidence across multi-event and target-absent queries.
Together, these gains indicate that policy--tool coevolution distills temporal grounding experience into a transferable skill for acquiring long-video evidence more accurately without updating model parameters.

\subsubsection{Inference Efficiency}
\label{sec:inference_efficiency}

Table~\ref{tab:efficiency} illustrates the improvements in inference efficiency achieved through policy--tool coevolution.
Average visual tokens per query drop from 202.6k to 141.9k (\textbf{30.0\%} fewer) on the VUE-LVTR held-out set, with reductions of \textbf{11.4\%} on ExtremeWhenBench and \textbf{18.9\%} on CoMET-Bench.
Model calls also decrease across all three benchmarks, suggesting that the VLM equipped with the evolved skill makes more efficient and targeted decisions about how to observe ultra-long videos.
A closer examination of the ExtremeWhenBench results reveals that tokens for image-based observations increase from 179.3k to 197.5k, whereas those for video-based observations fall from 47.8k to 3.7k, reflecting the reallocation of the visual budget during coevolution.

\subsubsection{Cross-VLM Generalization and Cross-Task Transfer}
\label{sec:generalization_transfer}

For different VLM backbones, as summarized in Table~\ref{tab:generalization}, policy--tool coevolution improves ExtremeWhenBench mIoU by \textbf{99.2\%} and \textbf{24.7\%} on Qwen3.5-9B~\citep{qwen2026qwen35} and Qwen3.6-27B~\citep{qwen2026qwen36}, respectively, which supports the framework's cross-VLM generalizability.
In addition, when the evolved skill is applied directly to general long-video QA, \mbox{Qwen3.5-27B}'s overall accuracy on LVBench and LSDBench improves by \textbf{8.65} and \textbf{7.98} points, respectively, demonstrating that the coevolved policies and tools can transfer effectively to broader long-video understanding scenarios.

\begin{table*}[t]
  \centering
  \providecolor{AcademicBlue}{RGB}{45,85,133}
\begin{minipage}[t]{\textwidth}
  \vspace{0pt}
  \centering
  \setlength{\abovecaptionskip}{0pt}
  \setlength{\belowcaptionskip}{6pt}
  \caption{\textbf{Generalization across VLMs and transfer to long-video QA.} \textbf{Bold}: results with the evolved skill; \textcolor{AcademicBlue}{blue}: gains over the base skill, with \textcolor{AcademicBlue}{$\uparrow$} indicating relative improvements.}
  \label{tab:generalization}
  \ExpMainTableStyle{3.5pt}
  \renewcommand{\arraystretch}{1.85}
  \resizebox{\linewidth}{!}{%
  \begin{tabular}{@{}>{\centering\arraybackslash}m{3.0cm}|l|*{3}{N{1.41cm}}|*{2}{N{1.41cm}}|*{2}{N{1.41cm}}|N{1.41cm}|N{1.41cm}@{}}
    \toprule[1.2pt]
      & & \multicolumn{3}{c|}{\textbf{VUE-LVTR}}
      & \multicolumn{2}{c|}{\textbf{ExtremeWhenBench}}
      & \multicolumn{2}{c|}{\textbf{CoMET-Bench}}
      & \multicolumn{1}{c|}{\textbf{LVBench}}
      & \multicolumn{1}{c}{\textbf{LSDBench}} \\
      & & \multicolumn{3}{c|}{\shortstack{\ExpTableSecondary 30.1--105.5 min\\[-1pt]\ExpTableSecondary (mean 50.9 min)}}
      & \multicolumn{2}{c|}{\shortstack{\ExpTableSecondary 45.0--542.5 min\\[-1pt]\ExpTableSecondary (mean 75.8 min)}}
      & \multicolumn{2}{c|}{\shortstack{\ExpTableSecondary 30.0--123.7 min\\[-1pt]\ExpTableSecondary (mean 50.5 min)}}
      & \multicolumn{1}{c|}{\ExpTableSecondary LVQA}
      & \multicolumn{1}{c}{\ExpTableSecondary LVQA} \\
    \cmidrule(lr){3-5}\cmidrule(lr){6-7}\cmidrule(lr){8-9}\cmidrule(lr){10-10}\cmidrule(l){11-11}
    \noalign{\vbox to 0pt{\vss\hbox{%
      \hskip\dimexpr3.0cm+\tabcolsep\relax
      \vrule width\arrayrulewidth
        height\dimexpr\aboverulesep+\cmidrulewidth+\belowrulesep\relax depth0pt}}}
      {\ExpTableLabel\textbf{VLM}} & {\ExpTableLabel\textbf{Method}}
      & \footnotesize\shortstack{\textbf{\ExpTableSecondary Precision}\\[-1pt]\textbf{\ExpTableSecondary AUC} $\uparrow$}
      & \footnotesize\shortstack{\textbf{\ExpTableSecondary Recall}\\[-1pt]\textbf{\ExpTableSecondary AUC} $\uparrow$}
      & \footnotesize\shortstack{\textbf{\ExpTableSecondary IoU}\\[-1pt]\textbf{\ExpTableSecondary AUC} $\uparrow$}
      & \footnotesize\shortstack{\textbf{\ExpTableSecondary Mean}\\[-1pt]\textbf{\ExpTableSecondary IoU} $\uparrow$}
      & \footnotesize\shortstack{\textbf{\ExpTableSecondary Recall}\\[-1pt]\textbf{\ExpTableSecondary @0.5} $\uparrow$}
      & \footnotesize\shortstack{\textbf{\ExpTableSecondary Mean}\\[-1pt]\textbf{\ExpTableSecondary IoU} $\uparrow$}
      & \footnotesize\shortstack{\textbf{\ExpTableSecondary Rejection}\\[-1pt]\textbf{\ExpTableSecondary F1} $\uparrow$}
      & \footnotesize\shortstack{\textbf{\ExpTableSecondary Overall}\\[-1pt]\textbf{\ExpTableSecondary Acc. (\%)} $\uparrow$}
      & \footnotesize\shortstack{\textbf{\ExpTableSecondary Overall}\\[-1pt]\textbf{\ExpTableSecondary Acc. (\%)} $\uparrow$} \\
    \midrule
    \cellcolor{white}\multirow{4}{=}{\centering\ExpTableLabel Qwen3.5-9B}
      & \cellcolor{gray!15}+ Base Skill
      & \cellcolor{gray!15}0.3240 & \cellcolor{gray!15}0.3425 & \cellcolor{gray!15}0.2405
      & \cellcolor{gray!15}0.0626 & \cellcolor{gray!15}0.0576
      & \cellcolor{gray!15}0.0672 & \cellcolor{gray!15}67.51
      & \cellcolor{gray!15}40.09 & \cellcolor{gray!15}49.08 \\
      & \textbf{+ Evolved Skill}
      & \textbf{0.4287} & \textbf{0.4707} & \textbf{0.3244}
      & \textbf{0.1247} & \textbf{0.1170}
      & \textbf{0.0852} & \textbf{68.80}
      & \textbf{45.84} & \textbf{53.68} \\
      & \textbf{Evolution Gain}
      & {\color{AcademicBlue}\textbf{+0.1047}}
      & {\color{AcademicBlue}\textbf{+0.1282}}
      & {\color{AcademicBlue}\textbf{+0.0839}}
      & {\color{AcademicBlue}\textbf{+0.0621}}
      & {\color{AcademicBlue}\textbf{+0.0594}}
      & {\color{AcademicBlue}\textbf{+0.0180}}
      & {\color{AcademicBlue}\textbf{+1.29}}
      & {\color{AcademicBlue}\textbf{+5.75}}
      & {\color{AcademicBlue}\textbf{+4.60}} \\[-5pt]
      &
      & {\color{AcademicBlue}\textbf{\ExpTableSecondary$\uparrow$32.3\%}}
      & {\color{AcademicBlue}\textbf{\ExpTableSecondary$\uparrow$37.4\%}}
      & {\color{AcademicBlue}\textbf{\ExpTableSecondary$\uparrow$34.9\%}}
      & {\color{AcademicBlue}\textbf{\ExpTableSecondary$\uparrow$99.2\%}}
      & {\color{AcademicBlue}\textbf{\ExpTableSecondary$\uparrow$103.1\%}}
      & {\color{AcademicBlue}\textbf{\ExpTableSecondary$\uparrow$26.8\%}}
      & {\color{AcademicBlue}\textbf{\ExpTableSecondary$\uparrow$1.9\%}}
      & {\color{AcademicBlue}\textbf{\ExpTableSecondary$\uparrow$14.3\%}}
      & {\color{AcademicBlue}\textbf{\ExpTableSecondary$\uparrow$9.4\%}} \\
    \midrule
    \cellcolor{white}\multirow{4}{=}{\centering\ExpTableLabel Qwen3.5-27B}
      & \cellcolor{gray!15}+ Base Skill
      & \cellcolor{gray!15}0.4706 & \cellcolor{gray!15}0.4789 & \cellcolor{gray!15}0.4107
      & \cellcolor{gray!15}0.1507 & \cellcolor{gray!15}0.1478
      & \cellcolor{gray!15}0.1355 & \cellcolor{gray!15}66.13
      & \cellcolor{gray!15}45.45 & \cellcolor{gray!15}61.27 \\
      & \textbf{+ Evolved Skill}
      & \textbf{0.5918} & \textbf{0.5773} & \textbf{0.5137}
      & \textbf{0.2636} & \textbf{0.2785}
      & \textbf{0.1635} & \textbf{77.06}
      & \textbf{54.10} & \textbf{69.25} \\
      & \textbf{Evolution Gain}
      & {\color{AcademicBlue}\textbf{+0.1212}}
      & {\color{AcademicBlue}\textbf{+0.0984}}
      & {\color{AcademicBlue}\textbf{+0.1030}}
      & {\color{AcademicBlue}\textbf{+0.1129}}
      & {\color{AcademicBlue}\textbf{+0.1307}}
      & {\color{AcademicBlue}\textbf{+0.0280}}
      & {\color{AcademicBlue}\textbf{+10.93}}
      & {\color{AcademicBlue}\textbf{+8.65}}
      & {\color{AcademicBlue}\textbf{+7.98}} \\[-5pt]
      &
      & {\color{AcademicBlue}\textbf{\ExpTableSecondary$\uparrow$25.8\%}}
      & {\color{AcademicBlue}\textbf{\ExpTableSecondary$\uparrow$20.5\%}}
      & {\color{AcademicBlue}\textbf{\ExpTableSecondary$\uparrow$25.1\%}}
      & {\color{AcademicBlue}\textbf{\ExpTableSecondary$\uparrow$74.9\%}}
      & {\color{AcademicBlue}\textbf{\ExpTableSecondary$\uparrow$88.4\%}}
      & {\color{AcademicBlue}\textbf{\ExpTableSecondary$\uparrow$20.7\%}}
      & {\color{AcademicBlue}\textbf{\ExpTableSecondary$\uparrow$16.5\%}}
      & {\color{AcademicBlue}\textbf{\ExpTableSecondary$\uparrow$19.0\%}}
      & {\color{AcademicBlue}\textbf{\ExpTableSecondary$\uparrow$13.0\%}} \\
    \midrule
    \cellcolor{white}\multirow{4}{=}{\centering\ExpTableLabel Qwen3.6-27B}
      & \cellcolor{gray!15}+ Base Skill
      & \cellcolor{gray!15}0.5098 & \cellcolor{gray!15}0.5465 & \cellcolor{gray!15}0.4506
      & \cellcolor{gray!15}0.1691 & \cellcolor{gray!15}0.1694
      & \cellcolor{gray!15}0.1312 & \cellcolor{gray!15}68.53
      & \cellcolor{gray!15}48.55 & \cellcolor{gray!15}61.12 \\
      & \textbf{+ Evolved Skill}
      & \textbf{0.5519} & \textbf{0.5727} & \textbf{0.4985}
      & \textbf{0.2109} & \textbf{0.2129}
      & \textbf{0.1477} & \textbf{72.37}
      & \textbf{54.36} & \textbf{65.64} \\
      & \textbf{Evolution Gain}
      & {\color{AcademicBlue}\textbf{+0.0421}}
      & {\color{AcademicBlue}\textbf{+0.0262}}
      & {\color{AcademicBlue}\textbf{+0.0479}}
      & {\color{AcademicBlue}\textbf{+0.0418}}
      & {\color{AcademicBlue}\textbf{+0.0435}}
      & {\color{AcademicBlue}\textbf{+0.0165}}
      & {\color{AcademicBlue}\textbf{+3.84}}
      & {\color{AcademicBlue}\textbf{+5.81}}
      & {\color{AcademicBlue}\textbf{+4.52}} \\[-5pt]
      &
      & {\color{AcademicBlue}\textbf{\ExpTableSecondary$\uparrow$8.3\%}}
      & {\color{AcademicBlue}\textbf{\ExpTableSecondary$\uparrow$4.8\%}}
      & {\color{AcademicBlue}\textbf{\ExpTableSecondary$\uparrow$10.6\%}}
      & {\color{AcademicBlue}\textbf{\ExpTableSecondary$\uparrow$24.7\%}}
      & {\color{AcademicBlue}\textbf{\ExpTableSecondary$\uparrow$25.7\%}}
      & {\color{AcademicBlue}\textbf{\ExpTableSecondary$\uparrow$12.6\%}}
      & {\color{AcademicBlue}\textbf{\ExpTableSecondary$\uparrow$5.6\%}}
      & {\color{AcademicBlue}\textbf{\ExpTableSecondary$\uparrow$12.0\%}}
      & {\color{AcademicBlue}\textbf{\ExpTableSecondary$\uparrow$7.4\%}} \\
    \bottomrule[1.2pt]
  \end{tabular}
  }
\end{minipage}

  \par\vspace{\floatsep}
  \centering
  \sbox{\ExpAblationFigureBox}{%
    \includegraphics[width=0.345\textwidth]{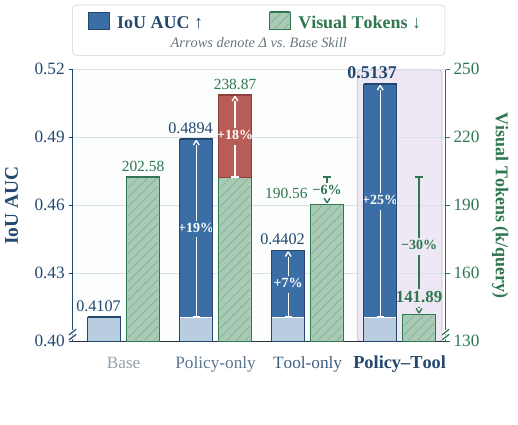}}
  \setbox\ExpAblationTableBox=\vbox{%
    \hsize=0.635\textwidth\linewidth=\hsize
    \centering\providecolor{AcademicBlue}{RGB}{45,85,133}
\providecolor{EfficiencyGreen}{RGB}{46,119,78}
\providecolor{AllocationPurple}{RGB}{117,87,150}

\ExpTableStyle{2.5pt}
\resizebox{\linewidth}{!}{%
    \begin{tabular}{@{}>{\centering\arraybackslash}m{0.8cm}>{\centering\arraybackslash}m{0.8cm}|l|*{4}{N{1.45cm}}N{1.94cm}@{}}
      \toprule[1.2pt]
      \shortstack{\textbf{\ExpTableSecondary Image}\\[-1pt]\textbf{\ExpTableSecondary Obs.}}
        & \footnotesize\shortstack{\textbf{\ExpTableSecondary Video}\\[-1pt]\textbf{\ExpTableSecondary Obs.}}
        & \textbf{Qwen3.5-27B}
        & \footnotesize\shortstack{\textbf{\ExpTableSecondary Precision}\\[-1pt]\textbf{\ExpTableSecondary AUC} $\uparrow$}
        & \footnotesize\shortstack{\textbf{\ExpTableSecondary Recall}\\[-1pt]\textbf{\ExpTableSecondary AUC} $\uparrow$}
        & \footnotesize\shortstack{\textbf{\ExpTableSecondary IoU}\\[-1pt]\textbf{\ExpTableSecondary AUC} $\uparrow$}
        & \footnotesize\shortstack{\textbf{\ExpTableSecondary Visual}\\[-1pt]\textbf{\ExpTableSecondary Tokens} $\downarrow$}
        & \footnotesize\shortstack{\textbf{\ExpTableSecondary Image / Video}\\[-1pt]\textbf{\ExpTableSecondary Tokens}} \\
      \midrule

      \cellcolor{white}\multirow{2}{*}{{\Large\ding{51}}}
        & \cellcolor{white}\multirow{2}{*}{{\Large\ding{55}}}
        & \cellcolor{gray!15}+ Base Skill
        & \cellcolor{gray!15}0.4903 & \cellcolor{gray!15}0.4698 & \cellcolor{gray!15}0.4173
        & \cellcolor{gray!15}185.49 & \cellcolor{gray!15}185.49 / \textemdash \\
        & & + Evolved Skill
        & 0.4917 & 0.5124 & 0.4374
        & 194.60 & 194.60 / \textemdash \\
      \midrule

      \cellcolor{white}\multirow{2}{*}{{\Large\ding{55}}}
        & \cellcolor{white}\multirow{2}{*}{{\Large\ding{51}}}
        & \cellcolor{gray!15}+ Base Skill
        & \cellcolor{gray!15}0.3474 & \cellcolor{gray!15}0.3092 & \cellcolor{gray!15}0.2457
        & \cellcolor{gray!15}693.52 & \cellcolor{gray!15}\textemdash{} / 693.52 \\
        & & + Evolved Skill
        & 0.5680 & 0.5653 & 0.4772
        & 221.98 & \textemdash{} / 221.98 \\
      \midrule

      \cellcolor{white}\multirow{4}{*}{{\Large\ding{51}}}
        & \cellcolor{white}\multirow{4}{*}{{\Large\ding{51}}}
        & \cellcolor{gray!15}+ Base Skill
        & \cellcolor{gray!15}0.4706 & \cellcolor{gray!15}0.4789 & \cellcolor{gray!15}0.4107
        & \cellcolor{gray!15}202.58 & \cellcolor{gray!15}162.01 / 40.57 \\
        & & \textbf{+ Evolved Skill}
        & \textbf{0.5918} & \textbf{0.5773} & \textbf{0.5137}
        & \textbf{141.89} & \textbf{141.37 / 0.52} \\
        & & \textbf{$\Delta$ vs. Image-only}
        & {\color{AcademicBlue}\textbf{+0.1001}}
        & {\color{AcademicBlue}\textbf{+0.0649}}
        & {\color{AcademicBlue}\textbf{+0.0763}}
        & {\color{EfficiencyGreen}\textbf{$-$52.71}}
        & {\color{EfficiencyGreen}\textbf{$-$53.23}} / \textemdash \\
        & & \textbf{$\Delta$ vs. Video-only}
        & {\color{AcademicBlue}\textbf{+0.0238}}
        & {\color{AcademicBlue}\textbf{+0.0120}}
        & {\color{AcademicBlue}\textbf{+0.0365}}
        & {\color{EfficiencyGreen}\textbf{$-$80.09}}
        & \textemdash{} / {\color{EfficiencyGreen}\textbf{$-$221.46}} \\
      \bottomrule[1.2pt]
\end{tabular}%
}
\par}
  \setlength{\ExpAblationPanelHeight}{%
    \dimexpr\ht\ExpAblationTableBox+\dp\ExpAblationTableBox\relax}
  \ifdim\dimexpr\ht\ExpAblationFigureBox+\dp\ExpAblationFigureBox\relax>\ExpAblationPanelHeight
    \setlength{\ExpAblationPanelHeight}{%
      \dimexpr\ht\ExpAblationFigureBox+\dp\ExpAblationFigureBox\relax}
  \fi
  \begin{minipage}[t]{0.345\textwidth}
    \vspace{0pt}
    \begingroup
  \centering
  \ifdim\ExpAblationPanelHeight>0pt
    \begin{minipage}[c][\ExpAblationPanelHeight][c]{\linewidth}
      \centering\usebox{\ExpAblationFigureBox}
    \end{minipage}\par
  \else
    \includegraphics[width=\linewidth]{figures/policy_tool_ablation_fig.pdf}\par
  \fi
  \captionof{figure}{\raggedright\textbf{Policy--tool coevolution ablation} on the VUE-LVTR held-out set. Arrows show relative changes from the base skill.}
  \label{fig:policy-tool-ablation}
\endgroup

  \end{minipage}%
  \hfill
  \begin{minipage}[t]{0.635\textwidth}
    \vspace{0pt}
    \begingroup
  \centering
  \ifdim\ExpAblationPanelHeight>0pt
    \begin{minipage}[c][\ExpAblationPanelHeight][c]{\linewidth}
      \centering\usebox{\ExpAblationTableBox}
    \end{minipage}\par
  \else
    {\par}
  \fi
  \captionof{table}{\raggedright\textbf{Image--video coordination ablation} on the VUE-LVTR held-out set. \textbf{Bold} marks results with the evolved image+video skill; \textcolor{AcademicBlue}{blue} and \textcolor{EfficiencyGreen}{green} denote its performance gains and reductions in visual token cost relative to evolved single-modality skills. Tokens: k/query.}
  \label{tab:image-video-ablation}
\endgroup

  \end{minipage}
\end{table*}

\subsection{Ablations and Analysis}
\label{sec:ablations_analysis}

\subsubsection{Ablation on Policy--Tool Coevolution}
\label{sec:policy_tool_ablation}

To assess the benefit of policy--tool coevolution and each component's contribution, we compare policy-only and tool-only variants that freeze the tools and policies during evolution, respectively.
Both use the same evolution set, evolution configuration, and VUE-LVTR held-out evaluation protocol as the full method.
As shown in Figure~\ref{fig:policy-tool-ablation}, both variants improve grounding accuracy over the base skill, confirming policies and tools as effective evolution targets.
Joint evolution achieves the best accuracy--cost combination, exceeding policy-only and tool-only by \textbf{0.0243} and \textbf{0.0735} in IoU AUC while reducing visual token cost by \textbf{40.6\%} and \textbf{25.5\%}, respectively.
This reveals that more efficient evidence acquisition from ultra-long videos relies on both executable media tools to expand the available observation capabilities and high-level policies to select and orchestrate them.
In particular, freezing the policies during evolution leads to a larger accuracy loss than freezing the tools, suggesting that stronger observation capabilities can yield substantial gains in ultra-long video temporal grounding when strategically organized into an efficient and accurate evidence search process.

\begin{figure*}[t]
  \centering
  \settoheight{\ExpEvolutionPanelHeight}{%
    \includegraphics[width=0.605\textwidth]{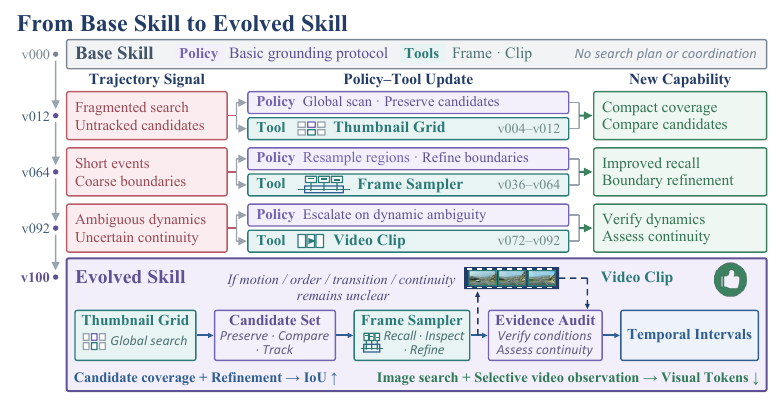}}
  \begin{minipage}[t]{0.375\textwidth}
    \vspace{0pt}
    \input{figures/evolution_dynamics}
  \end{minipage}%
  \hfill
  \begin{minipage}[t]{0.605\textwidth}
    \vspace{0pt}
    \input{figures/skill_evolution}
  \end{minipage}
\end{figure*}

\subsubsection{Ablation on Image--Video Coordination}
\label{sec:image_video_ablation}

We further conduct image-only and video-only ablations, each restricting the skill's observations to a single modality during evolution and evaluation, to isolate each modality's contribution and assess coordinated image--video observation.
As shown in Table~\ref{tab:image-video-ablation}, both single-modality variants improve grounding performance through evolution, and image--video coordination yields further benefits: the evolved image+video skill outperforms both single-modality skills across all grounding metrics while using fewer visual tokens.
These empirical results underscore that image-based and video-based observations are suited to different contexts in ultra-long video temporal grounding, and that strategically coordinating the two makes better use of their complementary strengths, achieving accuracy gains and cost reductions in tandem.

\subsubsection{Analysis of Skill Evolution}
\label{sec:skill_evolution_analysis}

Figure~\ref{fig:evolution-dynamics} illustrates online performance--cost dynamics during skill evolution.
As evolution proceeds, the skill achieves higher grounding accuracy with lower visual token cost more consistently.
Figure~\ref{fig:skill-evolution} further shows how candidate omissions, boundary errors, and dynamic ambiguities exposed in trajectories drive coordinated adjustments to observation capabilities and their orchestration: image-based tools with candidate-retention and local-refinement policies improve evidence coverage and boundary judgment, while video-based observation targets action verification, order discrimination, and continuity assessment.
Together, these advances suggest that policy--tool coevolution distills execution feedback into mutually adapted observation capabilities and policies, letting the VLM allocate its visual budget according to evidence needs for more accurate yet cheaper grounding.
Qualitative case studies are provided in Appendix~\ref{app:cases}.

\section{Conclusion}
\label{sec:conclusion}

We propose CoEvoWhen, a policy--tool coevolution framework that jointly evolves high-level policies and executable media tools from agentic reasoning trajectories into a reusable external skill for ultra-long video temporal grounding.
Equipped with the evolved skill, the VLM autonomously orchestrates tools under policy guidance, coordinating image-based and video-based observations without relying on a stronger external planner.
Comprehensive experiments show that policy--tool coevolution improves grounding accuracy while reducing visual token cost, with gains generalizing across VLMs and transferring to long-video QA without further task-specific evolution.
We believe that coevolving what a VLM can observe with how it decides to observe offers a promising path toward seeing less yet understanding more in ultra-long videos.

\bibliography{main}
\bibliographystyle{main}

\clearpage
\setcounter{topnumber}{2}
\setcounter{dbltopnumber}{2}
\appendix
\setcounter{topnumber}{3}
\setlength{\floatsep}{8pt plus 2pt minus 2pt}
\setlength{\intextsep}{14pt plus 2pt minus 2pt}
\section{Appendix Overview}
\label{app:overview}
The appendix is organized as follows:

\begingroup\footnotesize
\hypersetup{hidelinks}
\newcommand{\ovlink}[2]{\hyperref[#1]{\textcolor{AppendixLinkBlue}{\textbf{Section~\ref*{#1}:}}\ \textcolor{black}{\textbf{#2}}}}
\newcommand{\ovdesc}[1]{\begin{itemize}[label={},leftmargin=1.45em,topsep=0pt,itemsep=0pt,parsep=0pt,partopsep=0pt]\item #1\end{itemize}}
\newcommand{\ovitem}[2]{\item\ovlink{#1}{#2}}
\newcommand{\ovleaf}[3]{\item\ovlink{#1}{#2}\ovdesc{#3}}
\begin{itemize}[label=\footnotesize$\blacktriangleright$,leftmargin=1.5em,topsep=2pt,itemsep=1pt,parsep=0pt,partopsep=0pt]
  \ovitem{app:implementation}{Implementation Details}
  \begin{itemize}[label=\textbullet,leftmargin=1.25em,topsep=1pt,itemsep=1pt,parsep=0pt,partopsep=0pt]
    \ovleaf{app:evolution}{Details on Evolution}{Evolution set construction, skill initialization, and evolution settings.}
    \ovleaf{app:evaluation}{Details on Evaluation}{Data splits, evaluation metrics, inference settings, and baseline evaluation protocols.}
  \end{itemize}
  \ovitem{app:results}{More Results}
  \begin{itemize}[label=\textbullet,leftmargin=1.25em,topsep=1pt,itemsep=1pt,parsep=0pt,partopsep=0pt]
    \ovleaf{app:evolution-results}{Results on the Evolution Set}{Grounding results on the evolution set across three VLMs.}
    \ovleaf{app:qwen36-cost}{Visual Token Cost on Qwen3.6-27B}{Cumulative visual token cost and its image and video components.}
    \ovleaf{app:dimensions}{Results Across Multiple Dimensions}{Results across multiple query categories.}
  \end{itemize}
  \ovleaf{app:cases}{Case Study}{Two cases comparing execution trajectories with the base and evolved skills.}
  \ovleaf{app:limitations}{Limitations and Future Work}{Current limitations and future research directions.}
  \ovitem{app:prompts}{Prompts}
  \begin{itemize}[label=\textbullet,leftmargin=1.25em,topsep=1pt,itemsep=1pt,parsep=0pt,partopsep=0pt]
    \ovleaf{app:skill-prompts}{Skill Prompts}{System prompt and the base skill's \texttt{SKILL.md}.}
    \ovleaf{app:evaluation-prompts}{Evaluation Task Prompts}{Prompts for temporal grounding and long-video QA.}
    \ovleaf{app:updater-prompt}{Codex Updater Prompt}{Prompt for Codex to update policies and tools from execution feedback.}
  \end{itemize}
\end{itemize}
\endgroup
\vspace{0.15em}

\section{Implementation Details}
\label{app:implementation}

\subsection{Details on Evolution}
\label{app:evolution}

\subsubsection{Evolution Set Construction}
\label{app:evolution-set}

We select 100 queries from the purely visual temporal retrieval instances of VUE-TR-V2~\citep{vidi2026vidi25} to form the evolution set. Candidate instances are required to have an available source video of at least 30 minutes, spanning both long and ultra-long (at least 60 minutes) videos.

Selection jointly considers video duration, the number and length of target intervals, target sparsity, and whether the query involves action order, visual detail, or compound conditions. Table~\ref{tab:app-evolution-set} summarizes the main characteristics of the evolution set. Among these queries, 80 have sparse target events and 88 contain at least one target interval no longer than 10 seconds. Together, these instances allow the limited set of execution trajectories to cover observation needs ranging from long-range search to short-event localization and multi-candidate handling.

\begin{table}[!htbp]
\centering
\AppTableCaptionStyle
\caption{\textbf{Composition and difficulty characteristics of the evolution set.} The 100 queries come from 100 distinct videos. Sparse-target queries have a total target duration of at most 30 seconds, or at most 1\% of the video duration. Boundary-sensitive queries contain at least one target interval no longer than 10 seconds. Video-edge queries have targets within the first or last 30 seconds of the video.}
\label{tab:app-evolution-set}
\begingroup\AppTableStyle
\begin{tabularx}{\linewidth}{@{}>{\raggedright\arraybackslash}p{0.52\linewidth}|X}
\toprule
{\AppTableLabel\textbf{Property}} & {\AppTableLabel\textbf{Value}} \\
\midrule
Video Duration (Range / Mean) & {\AppTableFont 30.35--118.61 min / 52.90 min} \\
Keyword / Phrase / Sentence & {\AppTableFont 18 / 34 / 48} \\
Single-Interval / Multi-Interval & {\AppTableFont 56 / 44} \\
Sparse-Target Queries & {\AppTableFont 80} \\
Boundary-Sensitive Queries & {\AppTableFont 88} \\
Video-Edge Queries & {\AppTableFont 30} \\
\bottomrule
\end{tabularx}\endgroup
\end{table}

\subsubsection{Base Skill}
\label{app:base-skill}

The base skill $\mathcal S^{(0)}$ comprises policy documents, tool descriptions and interfaces, and the source code of the local media tools. The policies are organized into \texttt{SKILL.md} and three supporting documents: \texttt{policy-notes.md} carries guidance for task planning, \texttt{tool-selection.md} specifies how tools and observation modes are selected, and \texttt{tool-notes.md} records the input--output semantics and usage notes of each tool. Within this structure, the initial policies specify only a basic grounding protocol and elementary descriptions of the observation modes, without prescribing sophisticated strategies for global search, candidate management, or orchestration across modalities.

Table~\ref{tab:app-base-tools} lists the six tools provided by the base skill. Media preparation tools read source-video metadata or produce images and short video clips, which observation tools place in the same VLM's visual context. The \texttt{final\_answer} tool submits the final prediction. Each tool declares its name, description, and parameter interface in \texttt{tool.json}. Local media tools additionally provide an editable implementation in \texttt{tool.py}.

\begin{table}[!htbp]
\centering
\AppTableCaptionStyle
\caption{\textbf{Tools in the base skill and their functions.} Image and video observations are performed by the same VLM that executes the task. Local media tools prepare the inputs without calling additional visual models.}
\label{tab:app-base-tools}
\begingroup\AppTableStyle
\begin{tabularx}{\linewidth}{@{}>{\raggedright\arraybackslash}p{94pt}|>{\raggedright\arraybackslash}p{74pt}|X}
\toprule
{\AppTableLabel\textbf{Tool}} & {\AppTableLabel\textbf{Role}} & {\AppTableLabel\textbf{Function}} \\
\midrule
{\AppTableFont \texttt{probe\_media}} & Media Preparation & Read metadata such as video duration, resolution, and frame rate \\
{\AppTableFont \texttt{extract\_frames\_at}} & Media Preparation & Extract frames at specified source-video timestamps \\
{\AppTableFont \texttt{extract\_video\_clip}} & Media Preparation & Extract a video clip without audio from a specified time window \\
{\AppTableFont \texttt{inspect\_images}} & Image Observation & Add images to the VLM's visual context \\
{\AppTableFont \texttt{inspect\_video}} & Video Observation & Add video to the VLM's visual context \\
{\AppTableFont \texttt{final\_answer}} & Prediction & Submit the final answer in the format required by the current task \\
\bottomrule
\end{tabularx}\endgroup
\end{table}

\subsubsection{Evolution Settings}
\label{app:evolution-settings}

In the main experiments, we use Qwen3.5-27B~\citep{qwen2026qwen35} for temporal grounding and Codex (GPT-5.5, \texttt{xhigh})~\citep{openai2025codex,openai2026gpt55} as the external skill updater. Evolution makes a single pass over the 100 queries, forming a feedback batch after every four completed queries to update the policies and media tools. Table~\ref{tab:app-inference-settings} lists the VLM configuration and decoding settings. For the cross-VLM experiments, we evolve a separate skill on each of Qwen3.5-9B, Qwen3.5-27B, and Qwen3.6-27B~\citep{qwen2026qwen36}, and compare the base and evolved skills on the corresponding model. For the cross-task experiments, we directly apply the frozen skill evolved for grounding to long-video QA, without additional task-specific evolution.

\subsection{Details on Evaluation}
\label{app:evaluation}

\subsubsection{Data Splits}
\label{app:evaluation-sets}

Table~\ref{tab:app-evaluation-sets} summarizes the query counts of the three temporal grounding benchmarks and their splits for evolution and evaluation. VUE-LVTR is constructed from VUE-TR~\citep{vidi2025vidi} and VUE-TR-V2~\citep{vidi2026vidi25}, which contain 1,598 and 1,600 queries, respectively, of which 1,514 and 1,590 have their corresponding source videos available. We select purely visual queries whose corresponding video is at least 30 minutes long, remove 19 duplicate queries shared between the two sources and 6 queries with unavailable source videos, and obtain 407 VUE-LVTR queries, with VUE-TR and VUE-TR-V2 contributing 115 and 292 queries, respectively. Following the procedure described in Appendix~\ref{app:evolution-set}, we select 100 of the VUE-TR-V2 queries for skill evolution, and the remaining 307 queries, disjoint from the evolution set, form the held-out evaluation set.

We evaluate on the full official test set of ExtremeWhenBench~\citep{seo2026extremewhenbench}, which comprises 2,273 queries. For CoMET-Bench~\citep{zou2026comet}, among its 2,789 official queries, we retain those with a corresponding video of at least 30 minutes, yielding an evaluation set of 1,599 queries that covers both target-present and target-absent cases.

\begin{table}[!htbp]
\centering
\AppTableCaptionStyle
\caption{\textbf{Query counts and data splits for evolution and evaluation on the three temporal grounding benchmarks.}}
\label{tab:app-evaluation-sets}
\begingroup\AppTableStyle
\begin{tabularx}{\linewidth}{@{}>{\raggedright\arraybackslash}p{0.34\linewidth}|*{3}{Z}}
\toprule
{\AppTableLabel\textbf{Benchmark}} & {\AppTableLabel\textbf{Total queries}} & {\AppTableLabel\textbf{Evolution queries}} & {\AppTableLabel\textbf{Evaluation queries}} \\
\midrule
VUE-LVTR & 407 & 100 & 307 \\
ExtremeWhenBench & 2,273 & 0 & 2,273 \\
CoMET-Bench & 2,789 & 0 & 1,599 \\
\bottomrule
\end{tabularx}\endgroup
\end{table}

\subsubsection{Evaluation Metrics}
\label{app:metrics}

We follow the evaluation protocol of each benchmark. Let $N$ denote the number of queries in the evaluation set under consideration. Following the notation introduced in Section~\ref{sec:problem_formulation}, $\mathcal Y_i^\ast$ and $\widehat{\mathcal Y}_i$ denote the ground-truth and predicted interval sets, respectively, with $N_i$ and $\widehat N_i$ denoting their corresponding interval counts. Here, $\mathbf{1}[\cdot]$ denotes the indicator function.

\noindent\textbf{VUE-LVTR.} We follow the VUE-TR evaluation protocol~\citep{vidi2025vidi,vidi2026vidi25}. Before scoring, predicted start and end times are rounded down and up to integer seconds, respectively, and overlapping or adjacent intervals are then merged. Let $\widehat U_i$ and $U_i^\ast$ denote the time sets covered by the predicted and ground-truth intervals, respectively, and let $\mu(\cdot)$ denote the total duration of a time set. Temporal precision, recall, and IoU for each query are
\[
\operatorname{Prec}_i=
\frac{\mu(\widehat U_i\cap U_i^\ast)}{\mu(\widehat U_i)},
\quad
\operatorname{Rec}_i=
\frac{\mu(\widehat U_i\cap U_i^\ast)}{\mu(U_i^\ast)},
\quad
\operatorname{IoU}_i=
\frac{\mu(\widehat U_i\cap U_i^\ast)}{\mu(\widehat U_i\cup U_i^\ast)}.
\]
A score of zero is assigned whenever the corresponding denominator is zero. These three metrics respectively measure the precision of the predicted intervals, the coverage of the ground-truth event, and the temporal overlap between the two.

AUC is computed from the hit-rate curve over thresholds, using all queries in the evaluation set. For $m\in\{\operatorname{Prec},\operatorname{Rec}\}$, the hit rate at threshold $t$ is
\[
C_m(t)=\frac{1}{N}\sum_{i=1}^{N}\mathbf{1}[m_i\ge t].
\]
For IoU AUC, we follow the strict comparison in the official implementation:
$C_{\mathrm{IoU}}(t)=N^{-1}\sum_i\mathbf{1}[\operatorname{IoU}_i>t]$.
With $t_k=k/100$, trapezoidal integration over $[0,1]$ at a step size of 0.01 gives
\[
\operatorname{AUC}(m)=
0.01\sum_{k=0}^{99}
\frac{C_m(t_k)+C_m(t_{k+1})}{2}.
\]
Precision AUC, Recall AUC, and IoU AUC thus summarize the respective hit rates across the full threshold range. At a fixed threshold, IoU@x is defined as
\[
\operatorname{IoU@x}=
\frac{1}{N}\sum_{i=1}^{N}\mathbf{1}[\operatorname{IoU}_i\ge x],
\qquad x\in\{0.3,0.5,0.7\}.
\]
\noindent\textbf{ExtremeWhenBench.} Each query corresponds to a single target interval~\citep{seo2026extremewhenbench}. Temporal IoU is computed from the intersection and union of the predicted and ground-truth intervals and then aggregated as
\[
\operatorname{mIoU}=\frac{1}{N}\sum_{i=1}^{N}\operatorname{IoU}_i,
\qquad
\operatorname{Recall@x}=
\frac{1}{N}\sum_{i=1}^{N}\mathbf{1}[\operatorname{IoU}_i\ge x],
\]
where $x\in\{0.3,0.5,0.7\}$. Predictions that cannot be parsed as a valid single interval receive zero IoU and remain in the averages. Results for Action, Environment, Object, Reaction, and Scene report mIoU within each query category.

\noindent\textbf{CoMET-Bench.} This benchmark covers multi-event grounding, event counting, and queries with absent targets~\citep{zou2026comet}. Event counting is evaluated on all queries. Given the predicted event count $\widehat N_i$ and ground-truth count $N_i$, we compute the mean absolute error (MAE) and off-by-one accuracy (OBO), the fraction of queries whose count error is at most one:
\[
\operatorname{MAE}=\frac{1}{N}\sum_{i=1}^{N}|\widehat N_i-N_i|,
\qquad
\operatorname{OBO}=\frac{1}{N}\sum_{i=1}^{N}
\mathbf{1}[|\widehat N_i-N_i|\le1].
\]
Pearson correlation measures the correlation between the predicted and ground-truth event counts. Denoting their respective means by $\overline{\widehat N}$ and $\overline N$, we have
\[
\rho=
\frac{\sum_i(\widehat N_i-\overline{\widehat N})(N_i-\overline N)}
{\sqrt{\sum_i(\widehat N_i-\overline{\widehat N})^2}
 \sqrt{\sum_i(N_i-\overline N)^2}}.
\]
Grounding metrics are computed for positive queries, indexed by $\mathcal D_+=\{i:N_i>0\}$, and macro-averaged over this set. For each ground-truth interval $y_{in}^\ast$, let its highest IoU with any predicted interval be
\[
b_{in}=\max_{\widehat y\in\widehat{\mathcal Y}_i}
\operatorname{IoU}(\widehat y,y_{in}^\ast).
\]
We set $b_{in}=0$ when the prediction set is empty. The mIoU and Recall@0.5 for query $i$ are then
\[
\operatorname{mIoU}_i=\frac{1}{N_i}\sum_{n=1}^{N_i}b_{in},
\qquad
\text{Recall@0.5}_i=
\frac{1}{N_i}\sum_{n=1}^{N_i}\mathbf{1}[b_{in}\ge0.5].
\]
F1@0.5 uses one-to-one matching. Predicted intervals are processed in order, with each matched to the first unmatched ground-truth interval whose IoU is at least 0.5. If $\mathrm{TP}_i$ is the number of successful matches,
\[
\text{F1@0.5}_i=\frac{2\mathrm{TP}_i}{\widehat N_i+N_i}.
\]

For negative queries, correct rejection requires an empty predicted interval set. We denote the index set of queries with no ground-truth events by $\mathcal D_-=\{i:N_i=0\}$. The correct rejection rate $r_-$ and positive-query coverage are defined as
\[
r_-=\frac{1}{|\mathcal D_-|}
\sum_{i\in\mathcal D_-}\mathbf{1}[\widehat N_i=0],
\qquad
\operatorname{PosCoverage}=
\frac{1}{|\mathcal D_+|}
\sum_{i\in\mathcal D_+}\mathbf{1}[\widehat N_i>0].
\]
The false positive rate is $\operatorname{FPR}=1-r_-$. Rejection-F1 jointly measures correct rejection and positive-query coverage:
\[
\operatorname{RejF1}=
100\cdot
\frac{2r_-\operatorname{PosCoverage}}
{r_-+\operatorname{PosCoverage}}.
\]
This metric is set to zero when the denominator is zero. Following the official implementation, $r_-$ is rounded to four decimal places before computing Rejection-F1. Rejection-F1 is reported on a 0--100 scale, while the other proportion-based metrics use a 0--1 scale.

\noindent\textbf{Visual token cost.} For query $i$, let $C_i$ denote the number of model calls. In call $c$, the model actually receives $T_{i,c}^{\mathrm{img}}$ image tokens and $T_{i,c}^{\mathrm{vid}}$ video tokens. The average cumulative visual cost is defined as
\[
\overline T^{m}=
\frac{1}{N}\sum_{i=1}^{N}\sum_{c=1}^{C_i}T_{i,c}^{m},
\quad m\in\{\mathrm{img},\mathrm{vid}\},
\qquad
\overline T^{\mathrm{vis}}=
\overline T^{\mathrm{img}}+\overline T^{\mathrm{vid}}.
\]
When an image or video re-enters a later request as part of the interaction history, its visual tokens are counted again for that request. In the main experiments, token counts are computed from the actual resizing and sampling parameters of each request, following the grid used by the Qwen3.5-27B visual processor, rather than simply counting media files. All queries, including those without a valid final prediction, contribute to the denominator of the average. Media that are generated but never passed to the model incur no visual token cost.

Interaction statistics are likewise averaged per query, with separate counts recorded for model calls, local tool calls, and image and video observations. The total number of visual observations is the sum of the image and video observation counts.

\subsubsection{Inference Settings}
\label{app:inference-settings}

The base and evolved skills remain fixed throughout evaluation. For a given VLM, both use the same set of queries, task prompts, decoding configuration, interaction budget, and scorer. Policy documents are incorporated into the model's system context, and tools are registered as callable functions.

Table~\ref{tab:app-inference-settings} summarizes the model configuration for the main Qwen3.5-27B experiments. Task execution during evolution uses the same VLM configuration. Image and video observations are not subject to separate call limits but share an interaction budget of at most 24 inference rounds per query.

\begin{table}[!htbp]
\centering
\AppTableCaptionStyle
\caption{\textbf{Model configuration for Qwen3.5-27B.} The base and evolved skills use identical runtime settings.}
\label{tab:app-inference-settings}
\begingroup\AppTableStyle
\begin{tabularx}{\linewidth}{@{}>{\raggedright\arraybackslash}p{0.52\linewidth}|X}
\toprule
{\AppTableLabel\textbf{Parameter}} & {\AppTableLabel\textbf{Setting}} \\
\midrule
Serving Engine & vLLM \\
Numerical Precision & bfloat16 \\
Maximum Context Length & 262,144 tokens \\
Thinking Mode & Enabled \\
Decoding & Greedy \\
Temperature / Top-p / Top-k & 0 / 1.0 / 0 \\
Maximum Generated Tokens per Call & 4,096 \\
\bottomrule
\end{tabularx}\endgroup
\end{table}

All experiments use visual inputs only, without audio, speech recognition, or transcripts. Predictions for VUE-LVTR and CoMET-Bench are sets of source-video time intervals, whereas ExtremeWhenBench requires a single interval. When transferring to LVBench~\citep{wang2025lvbench} and LSDBench~\citep{qu2025lsdbench}, the model continues to search for evidence using the policies and tools evolved for grounding. Task adaptation only supplies the QA instructions and answer options while changing the final output to a single option letter, and the skill is not updated on QA data.

\subsubsection{Baseline Evaluation Protocols}
\label{app:baselines}

We evaluate seven VLM-based and agent-based baselines on the three ultra-long video temporal grounding benchmarks. Among open-source VLMs, Qwen3.5-27B~\citep{qwen2026qwen35} receives video input with 768 frames uniformly sampled across the full duration, InternVL3.5-8B~\citep{wang2025internvl35} uses a 128-frame image sequence, and TimeLens-7B~\citep{zhang2026timelens} uses 384 video frames with textual timestamps. For the closed-source models Gemini 2.5 Flash~\citep{gemini2025gemini25} and GPT-5.6 Luna~\citep{openai2026gpt56}, the input consists of 128 uniformly sampled frames arranged chronologically in timestamped image grids. Agent-based baselines include VideoMind-7B~\citep{liu2026videomind}, in which grounding and verification roles collaborate for multi-step video reasoning, and EvoGround-7B~\citep{jung2026evoground}, which uses feedback between its Proposer and Solver for self-evolving training to improve temporal grounding.

\section{More Results}
\label{app:results}

\subsection{Results on the Evolution Set}
\label{app:evolution-results}

Table~\ref{tab:app-evolution-results} reports the results of applying the fixed base and evolved skills to the 100 evolution queries after evolution has completed. All reported metrics improve across the three VLMs, with IoU AUC gains of \textbf{0.0882} for Qwen3.5-9B, \textbf{0.0651} for Qwen3.5-27B, and \textbf{0.0523} for Qwen3.6-27B. These results characterize how well each skill fits the evolution data.

\begin{table}[!htb]
\centering
\AppTableCaptionStyle
\caption{\textbf{Grounding gains from policy--tool coevolution on the evolution set.} Results are obtained by evaluating the fixed skills after evolution. \textbf{Bold}: better results; \textcolor{AcademicBlue}{blue}: gains over the base skill. Evolution Gain reports absolute changes, with arrows indicating relative changes.}
\label{tab:app-evolution-results}
\begingroup\AppTableStyle
\begin{tabularx}{\linewidth}{@{}Q{72pt}|>{\raggedright\arraybackslash}m{72pt}|*{3}{Z}|*{3}{Z}}
\toprule
{\AppTableLabel\textbf{VLM}} & {\AppTableLabel\textbf{Method}} & \shortstack[c]{{\bfseries \AppTableSecondary Precision}\\{\bfseries \AppTableSecondary AUC $\uparrow$}} & \shortstack[c]{{\bfseries \AppTableSecondary Recall}\\{\bfseries \AppTableSecondary AUC $\uparrow$}} & \shortstack[c]{{\bfseries \AppTableSecondary IoU}\\{\bfseries \AppTableSecondary AUC $\uparrow$}} & {\bfseries \AppTableSecondary IoU@0.3 $\uparrow$} & {\bfseries \AppTableSecondary IoU@0.5 $\uparrow$} & {\bfseries \AppTableSecondary IoU@0.7 $\uparrow$} \\ \midrule
\cellcolor{white}\multirow{4}{=}{\centering\AppTableLabel Qwen3.5-9B} & \cellcolor{gray!15}+ Base Skill & \cellcolor{gray!15}{\AppTableFont 0.1759} & \cellcolor{gray!15}{\AppTableFont 0.1961} & \cellcolor{gray!15}{\AppTableFont 0.1142} & \cellcolor{gray!15}{\AppTableFont 0.1600} & \cellcolor{gray!15}{\AppTableFont 0.1300} & \cellcolor{gray!15}{\AppTableFont 0.0500} \\
 & \textbf{+ Evolved Skill} & {\textbf{\AppTableFont 0.3065}} & {\textbf{\AppTableFont 0.3545}} & {\textbf{\AppTableFont 0.2024}} & {\textbf{\AppTableFont 0.3000}} & {\textbf{\AppTableFont 0.2000}} & {\textbf{\AppTableFont 0.1100}} \\
 & \textbf{Evolution Gain} & {\color{AcademicBlue}\textbf{\AppTableFont +0.1306}} & {\color{AcademicBlue}\textbf{\AppTableFont +0.1584}} & {\color{AcademicBlue}\textbf{\AppTableFont +0.0882}} & {\color{AcademicBlue}\textbf{\AppTableFont +0.1400}} & {\color{AcademicBlue}\textbf{\AppTableFont +0.0700}} & {\color{AcademicBlue}\textbf{\AppTableFont +0.0600}} \\[-3.5pt]
 &  & {\color{AcademicBlue}\textbf{\AppTableSecondary$\uparrow$74.2\%}} & {\color{AcademicBlue}\textbf{\AppTableSecondary$\uparrow$80.8\%}} & {\color{AcademicBlue}\textbf{\AppTableSecondary$\uparrow$77.2\%}} & {\color{AcademicBlue}\textbf{\AppTableSecondary$\uparrow$87.5\%}} & {\color{AcademicBlue}\textbf{\AppTableSecondary$\uparrow$53.8\%}} & {\color{AcademicBlue}\textbf{\AppTableSecondary$\uparrow$120.0\%}} \\
\midrule
\cellcolor{white}\multirow{4}{=}{\centering\AppTableLabel Qwen3.5-27B} & \cellcolor{gray!15}+ Base Skill & \cellcolor{gray!15}{\AppTableFont 0.3620} & \cellcolor{gray!15}{\AppTableFont 0.3047} & \cellcolor{gray!15}{\AppTableFont 0.2556} & \cellcolor{gray!15}{\AppTableFont 0.3400} & \cellcolor{gray!15}{\AppTableFont 0.2800} & \cellcolor{gray!15}{\AppTableFont 0.1700} \\
 & \textbf{+ Evolved Skill} & {\textbf{\AppTableFont 0.4285}} & {\textbf{\AppTableFont 0.3916}} & {\textbf{\AppTableFont 0.3207}} & {\textbf{\AppTableFont 0.4300}} & {\textbf{\AppTableFont 0.3500}} & {\textbf{\AppTableFont 0.2400}} \\
 & \textbf{Evolution Gain} & {\color{AcademicBlue}\textbf{\AppTableFont +0.0665}} & {\color{AcademicBlue}\textbf{\AppTableFont +0.0869}} & {\color{AcademicBlue}\textbf{\AppTableFont +0.0651}} & {\color{AcademicBlue}\textbf{\AppTableFont +0.0900}} & {\color{AcademicBlue}\textbf{\AppTableFont +0.0700}} & {\color{AcademicBlue}\textbf{\AppTableFont +0.0700}} \\[-3.5pt]
 &  & {\color{AcademicBlue}\textbf{\AppTableSecondary$\uparrow$18.4\%}} & {\color{AcademicBlue}\textbf{\AppTableSecondary$\uparrow$28.5\%}} & {\color{AcademicBlue}\textbf{\AppTableSecondary$\uparrow$25.5\%}} & {\color{AcademicBlue}\textbf{\AppTableSecondary$\uparrow$26.5\%}} & {\color{AcademicBlue}\textbf{\AppTableSecondary$\uparrow$25.0\%}} & {\color{AcademicBlue}\textbf{\AppTableSecondary$\uparrow$41.2\%}} \\
\midrule
\cellcolor{white}\multirow{4}{=}{\centering\AppTableLabel Qwen3.6-27B} & \cellcolor{gray!15}+ Base Skill & \cellcolor{gray!15}{\AppTableFont 0.3341} & \cellcolor{gray!15}{\AppTableFont 0.3307} & \cellcolor{gray!15}{\AppTableFont 0.2459} & \cellcolor{gray!15}{\AppTableFont 0.3600} & \cellcolor{gray!15}{\AppTableFont 0.2800} & \cellcolor{gray!15}{\AppTableFont 0.1400} \\
 & \textbf{+ Evolved Skill} & {\textbf{\AppTableFont 0.3732}} & {\textbf{\AppTableFont 0.4137}} & {\textbf{\AppTableFont 0.2982}} & {\textbf{\AppTableFont 0.4100}} & {\textbf{\AppTableFont 0.3300}} & {\textbf{\AppTableFont 0.2200}} \\
 & \textbf{Evolution Gain} & {\color{AcademicBlue}\textbf{\AppTableFont +0.0391}} & {\color{AcademicBlue}\textbf{\AppTableFont +0.0830}} & {\color{AcademicBlue}\textbf{\AppTableFont +0.0523}} & {\color{AcademicBlue}\textbf{\AppTableFont +0.0500}} & {\color{AcademicBlue}\textbf{\AppTableFont +0.0500}} & {\color{AcademicBlue}\textbf{\AppTableFont +0.0800}} \\[-3.5pt]
 &  & {\color{AcademicBlue}\textbf{\AppTableSecondary$\uparrow$11.7\%}} & {\color{AcademicBlue}\textbf{\AppTableSecondary$\uparrow$25.1\%}} & {\color{AcademicBlue}\textbf{\AppTableSecondary$\uparrow$21.3\%}} & {\color{AcademicBlue}\textbf{\AppTableSecondary$\uparrow$13.9\%}} & {\color{AcademicBlue}\textbf{\AppTableSecondary$\uparrow$17.9\%}} & {\color{AcademicBlue}\textbf{\AppTableSecondary$\uparrow$57.1\%}} \\
\bottomrule
\end{tabularx}
\endgroup
\end{table}

\subsection{Visual Token Cost on Qwen3.6-27B}
\label{app:qwen36-cost}

Table~\ref{tab:app-qwen36-cost} reports visual token cost for Qwen3.6-27B. After evolution, the average cumulative visual token cost decreases by \textbf{42.2\%} on the VUE-LVTR held-out set, \textbf{28.1\%} on ExtremeWhenBench, and \textbf{15.8\%} on CoMET-Bench. On CoMET-Bench, image tokens increase from 83.60k to 100.19k while video tokens decrease from 55.36k to 16.82k. The reduction in total cost is thus accompanied by a reallocation of the visual budget between the two observation modalities.

\begin{table}[!htb]
\centering
\AppTableCaptionStyle
\caption{\textbf{Visual token cost of the base and evolved skills on Qwen3.6-27B.} \textbf{Bold}: lower cost. \textcolor{EfficiencyGreen}{Green ($\downarrow$)} and \textcolor{AllocationPurple}{purple ($\uparrow$)} indicate decreases and increases relative to the base skill. $\Delta$ denotes evolved cost minus base cost, with arrows indicating relative changes. Tokens: k/query.}
\label{tab:app-qwen36-cost}
\begingroup\AppTableStyle
\begin{tabularx}{\linewidth}{@{}Q{72pt}|>{\raggedright\arraybackslash}m{72pt}|*{3}{Z}}
\toprule
{\AppTableLabel\textbf{Benchmark}} & {\AppTableLabel\textbf{Method}} & {\bfseries \AppTableSecondary Visual Tokens $\downarrow$} & {\bfseries \AppTableSecondary Image Tokens} & {\bfseries \AppTableSecondary Video Tokens} \\ \midrule
\cellcolor{white}\multirow{4}{=}{\centering\AppTableLabel VUE-LVTR\\[1pt]{\AppTableSecondary held-out}} & \cellcolor{gray!15}Base Skill & \cellcolor{gray!15}{\AppTableFont 241.21} & \cellcolor{gray!15}{\AppTableFont 214.98} & \cellcolor{gray!15}{\AppTableFont 26.23} \\
 & \textbf{Evolved Skill} & {\textbf{\AppTableFont 139.36}} & {\textbf{\AppTableFont 134.10}} & {\textbf{\AppTableFont 5.26}} \\
 & \textbf{$\Delta$ (Evolved $-$ Base)} & {\color{EfficiencyGreen}\textbf{\AppTableFont $-$101.85}} & {\color{EfficiencyGreen}\textbf{\AppTableFont $-$80.88}} & {\color{EfficiencyGreen}\textbf{\AppTableFont $-$20.97}} \\[-3.5pt]
 &  & {\color{EfficiencyGreen}\textbf{\AppTableSecondary$\downarrow$42.2\%}} & {\color{EfficiencyGreen}\textbf{\AppTableSecondary$\downarrow$37.6\%}} & {\color{EfficiencyGreen}\textbf{\AppTableSecondary$\downarrow$79.9\%}} \\
\midrule
\cellcolor{white}\multirow{4}{=}{\centering\AppTableLabel ExtremeWhenBench} & \cellcolor{gray!15}Base Skill & \cellcolor{gray!15}{\AppTableFont 220.29} & \cellcolor{gray!15}{\AppTableFont 195.24} & \cellcolor{gray!15}{\AppTableFont 25.05} \\
 & \textbf{Evolved Skill} & {\textbf{\AppTableFont 158.33}} & {\textbf{\AppTableFont 152.64}} & {\textbf{\AppTableFont 5.69}} \\
 & \textbf{$\Delta$ (Evolved $-$ Base)} & {\color{EfficiencyGreen}\textbf{\AppTableFont $-$61.96}} & {\color{EfficiencyGreen}\textbf{\AppTableFont $-$42.60}} & {\color{EfficiencyGreen}\textbf{\AppTableFont $-$19.36}} \\[-3.5pt]
 &  & {\color{EfficiencyGreen}\textbf{\AppTableSecondary$\downarrow$28.1\%}} & {\color{EfficiencyGreen}\textbf{\AppTableSecondary$\downarrow$21.8\%}} & {\color{EfficiencyGreen}\textbf{\AppTableSecondary$\downarrow$77.3\%}} \\
\midrule
\cellcolor{white}\multirow{4}{=}{\centering\AppTableLabel CoMET-Bench} & \cellcolor{gray!15}Base Skill & \cellcolor{gray!15}{\AppTableFont 138.96} & \cellcolor{gray!15}{\textbf{\AppTableFont 83.60}} & \cellcolor{gray!15}{\AppTableFont 55.36} \\
 & \textbf{Evolved Skill} & {\textbf{\AppTableFont 117.01}} & {\AppTableFont 100.19} & {\textbf{\AppTableFont 16.82}} \\
 & \textbf{$\Delta$ (Evolved $-$ Base)} & {\color{EfficiencyGreen}\textbf{\AppTableFont $-$21.95}} & {\color{AllocationPurple}\AppTableFont +16.59} & {\color{EfficiencyGreen}\textbf{\AppTableFont $-$38.54}} \\[-3.5pt]
 &  & {\color{EfficiencyGreen}\textbf{\AppTableSecondary$\downarrow$15.8\%}} & {\color{AllocationPurple}\AppTableSecondary$\uparrow$19.8\%} & {\color{EfficiencyGreen}\textbf{\AppTableSecondary$\downarrow$69.6\%}} \\
\bottomrule
\end{tabularx}
\endgroup
\end{table}

\subsection{Results Across Multiple Dimensions}
\label{app:dimensions}

\noindent\textbf{ExtremeWhenBench.} Table~\ref{tab:app-ewb-categories} reports mIoU by query category. Qwen3.5-9B and Qwen3.5-27B improve across all five categories, while Qwen3.6-27B improves on Action, Object, Reaction, and Scene, with a slight decrease on Environment. This variation across categories indicates that the distribution of evolution gains depends on both the model and the query type.

\begin{table}[!t]
\centering
\AppTableCaptionStyle
\caption{\textbf{Grounding results by query category on ExtremeWhenBench.} Each column reports mIoU ($\uparrow$) for the corresponding category. \textbf{Bold}: better results; \textcolor{AcademicBlue}{blue}: gains over the base skill. Evolution Gain reports absolute changes, with arrows indicating relative changes.}
\label{tab:app-ewb-categories}
\begingroup\AppTableStyle
\begin{tabularx}{\linewidth}{@{}Q{72pt}|>{\raggedright\arraybackslash}m{72pt}|*{5}{Z}}
\toprule
{\AppTableLabel\textbf{VLM}} & {\AppTableLabel\textbf{Method}} & {\bfseries \AppTableSecondary Action} & {\bfseries \AppTableSecondary Environment} & {\bfseries \AppTableSecondary Object} & {\bfseries \AppTableSecondary Reaction} & {\bfseries \AppTableSecondary Scene} \\ \midrule
\cellcolor{white}\multirow{4}{=}{\centering\AppTableLabel Qwen3.5-9B} & \cellcolor{gray!15}+ Base Skill & \cellcolor{gray!15}{\AppTableFont 0.0543} & \cellcolor{gray!15}{\AppTableFont 0.0986} & \cellcolor{gray!15}{\AppTableFont 0.0877} & \cellcolor{gray!15}{\AppTableFont 0.0439} & \cellcolor{gray!15}{\AppTableFont 0.0570} \\
 & \textbf{+ Evolved Skill} & {\textbf{\AppTableFont 0.1092}} & {\textbf{\AppTableFont 0.1485}} & {\textbf{\AppTableFont 0.1518}} & {\textbf{\AppTableFont 0.1218}} & {\textbf{\AppTableFont 0.1246}} \\
 & \textbf{Evolution Gain} & {\color{AcademicBlue}\textbf{\AppTableFont +0.0549}} & {\color{AcademicBlue}\textbf{\AppTableFont +0.0499}} & {\color{AcademicBlue}\textbf{\AppTableFont +0.0641}} & {\color{AcademicBlue}\textbf{\AppTableFont +0.0779}} & {\color{AcademicBlue}\textbf{\AppTableFont +0.0676}} \\[-3.5pt]
 &  & {\color{AcademicBlue}\textbf{\AppTableSecondary$\uparrow$101.1\%}} & {\color{AcademicBlue}\textbf{\AppTableSecondary$\uparrow$50.6\%}} & {\color{AcademicBlue}\textbf{\AppTableSecondary$\uparrow$73.1\%}} & {\color{AcademicBlue}\textbf{\AppTableSecondary$\uparrow$177.4\%}} & {\color{AcademicBlue}\textbf{\AppTableSecondary$\uparrow$118.6\%}} \\
\midrule
\cellcolor{white}\multirow{4}{=}{\centering\AppTableLabel Qwen3.5-27B} & \cellcolor{gray!15}+ Base Skill & \cellcolor{gray!15}{\AppTableFont 0.1446} & \cellcolor{gray!15}{\AppTableFont 0.2040} & \cellcolor{gray!15}{\AppTableFont 0.1862} & \cellcolor{gray!15}{\AppTableFont 0.1167} & \cellcolor{gray!15}{\AppTableFont 0.1377} \\
 & \textbf{+ Evolved Skill} & {\textbf{\AppTableFont 0.2459}} & {\textbf{\AppTableFont 0.2884}} & {\textbf{\AppTableFont 0.3427}} & {\textbf{\AppTableFont 0.1872}} & {\textbf{\AppTableFont 0.2664}} \\
 & \textbf{Evolution Gain} & {\color{AcademicBlue}\textbf{\AppTableFont +0.1013}} & {\color{AcademicBlue}\textbf{\AppTableFont +0.0844}} & {\color{AcademicBlue}\textbf{\AppTableFont +0.1565}} & {\color{AcademicBlue}\textbf{\AppTableFont +0.0705}} & {\color{AcademicBlue}\textbf{\AppTableFont +0.1287}} \\[-3.5pt]
 &  & {\color{AcademicBlue}\textbf{\AppTableSecondary$\uparrow$70.1\%}} & {\color{AcademicBlue}\textbf{\AppTableSecondary$\uparrow$41.4\%}} & {\color{AcademicBlue}\textbf{\AppTableSecondary$\uparrow$84.0\%}} & {\color{AcademicBlue}\textbf{\AppTableSecondary$\uparrow$60.4\%}} & {\color{AcademicBlue}\textbf{\AppTableSecondary$\uparrow$93.5\%}} \\
\midrule
\cellcolor{white}\multirow{4}{=}{\centering\AppTableLabel Qwen3.6-27B} & \cellcolor{gray!15}+ Base Skill & \cellcolor{gray!15}{\AppTableFont 0.1565} & \cellcolor{gray!15}{\textbf{\AppTableFont 0.2162}} & \cellcolor{gray!15}{\AppTableFont 0.2289} & \cellcolor{gray!15}{\AppTableFont 0.1259} & \cellcolor{gray!15}{\AppTableFont 0.1583} \\
 & \textbf{+ Evolved Skill} & {\textbf{\AppTableFont 0.2122}} & {\AppTableFont 0.2004} & {\textbf{\AppTableFont 0.2797}} & {\textbf{\AppTableFont 0.1681}} & {\textbf{\AppTableFont 0.1975}} \\
 & \textbf{Evolution Gain} & {\color{AcademicBlue}\textbf{\AppTableFont +0.0557}} & {\AppTableFont $-$0.0158} & {\color{AcademicBlue}\textbf{\AppTableFont +0.0508}} & {\color{AcademicBlue}\textbf{\AppTableFont +0.0422}} & {\color{AcademicBlue}\textbf{\AppTableFont +0.0392}} \\[-3.5pt]
 &  & {\color{AcademicBlue}\textbf{\AppTableSecondary$\uparrow$35.6\%}} & {\AppTableSecondary$\downarrow$7.3\%} & {\color{AcademicBlue}\textbf{\AppTableSecondary$\uparrow$22.2\%}} & {\color{AcademicBlue}\textbf{\AppTableSecondary$\uparrow$33.5\%}} & {\color{AcademicBlue}\textbf{\AppTableSecondary$\uparrow$24.8\%}} \\
\bottomrule
\end{tabularx}
\endgroup
\vspace{28pt}
\end{table}

\begin{table}[!t]
\centering
\AppTableCaptionStyle
\caption{\textbf{Event counting results on CoMET-Bench.} \textbf{Bold}: better results; \textcolor{AcademicBlue}{blue}: gains over the base skill. Evolution Gain reports absolute changes, with arrows indicating relative changes.}
\label{tab:app-comet-counting}
\begingroup\AppTableStyle
\begin{tabularx}{\linewidth}{@{}Q{72pt}|>{\raggedright\arraybackslash}m{72pt}|*{3}{Z}}
\toprule
{\AppTableLabel\textbf{VLM}} & {\AppTableLabel\textbf{Method}} & {\bfseries \AppTableSecondary MAE $\downarrow$} & {\bfseries \AppTableSecondary OBO $\uparrow$} & {\bfseries \AppTableSecondary Pearson $\uparrow$} \\ \midrule
\cellcolor{white}\multirow{4}{=}{\centering\AppTableLabel Qwen3.5-9B} & \cellcolor{gray!15}+ Base Skill & \cellcolor{gray!15}{\AppTableFont 3.7104} & \cellcolor{gray!15}{\AppTableFont 0.5378} & \cellcolor{gray!15}{\AppTableFont 0.3305} \\
 & \textbf{+ Evolved Skill} & {\textbf{\AppTableFont 3.6929}} & {\textbf{\AppTableFont 0.5516}} & {\textbf{\AppTableFont 0.6614}} \\
 & \textbf{Evolution Gain} & {\color{AcademicBlue}\textbf{\AppTableFont $-$0.0175}} & {\color{AcademicBlue}\textbf{\AppTableFont +0.0138}} & {\color{AcademicBlue}\textbf{\AppTableFont +0.3309}} \\[-3.5pt]
 &  & {\color{AcademicBlue}\textbf{\AppTableSecondary$\downarrow$0.5\%}} & {\color{AcademicBlue}\textbf{\AppTableSecondary$\uparrow$2.6\%}} & {\color{AcademicBlue}\textbf{\AppTableSecondary$\uparrow$100.1\%}} \\
\midrule
\cellcolor{white}\multirow{4}{=}{\centering\AppTableLabel Qwen3.5-27B} & \cellcolor{gray!15}+ Base Skill & \cellcolor{gray!15}{\AppTableFont 3.6717} & \cellcolor{gray!15}{\AppTableFont 0.5760} & \cellcolor{gray!15}{\AppTableFont 0.0775} \\
 & \textbf{+ Evolved Skill} & {\textbf{\AppTableFont 3.4290}} & {\textbf{\AppTableFont 0.6116}} & {\textbf{\AppTableFont 0.1258}} \\
 & \textbf{Evolution Gain} & {\color{AcademicBlue}\textbf{\AppTableFont $-$0.2427}} & {\color{AcademicBlue}\textbf{\AppTableFont +0.0356}} & {\color{AcademicBlue}\textbf{\AppTableFont +0.0483}} \\[-3.5pt]
 &  & {\color{AcademicBlue}\textbf{\AppTableSecondary$\downarrow$6.6\%}} & {\color{AcademicBlue}\textbf{\AppTableSecondary$\uparrow$6.2\%}} & {\color{AcademicBlue}\textbf{\AppTableSecondary$\uparrow$62.3\%}} \\
\midrule
\cellcolor{white}\multirow{4}{=}{\centering\AppTableLabel Qwen3.6-27B} & \cellcolor{gray!15}+ Base Skill & \cellcolor{gray!15}{\AppTableFont 3.6473} & \cellcolor{gray!15}{\AppTableFont 0.5716} & \cellcolor{gray!15}{\AppTableFont 0.0864} \\
 & \textbf{+ Evolved Skill} & {\textbf{\AppTableFont 3.5616}} & {\textbf{\AppTableFont 0.5816}} & {\textbf{\AppTableFont 0.1214}} \\
 & \textbf{Evolution Gain} & {\color{AcademicBlue}\textbf{\AppTableFont $-$0.0857}} & {\color{AcademicBlue}\textbf{\AppTableFont +0.0100}} & {\color{AcademicBlue}\textbf{\AppTableFont +0.0350}} \\[-3.5pt]
 &  & {\color{AcademicBlue}\textbf{\AppTableSecondary$\downarrow$2.3\%}} & {\color{AcademicBlue}\textbf{\AppTableSecondary$\uparrow$1.7\%}} & {\color{AcademicBlue}\textbf{\AppTableSecondary$\uparrow$40.5\%}} \\
\bottomrule
\end{tabularx}
\endgroup
\vspace{28pt}
\end{table}

\noindent\textbf{CoMET-Bench.} We examine the evolution gains in terms of event counting, rejection behavior on target-absent queries, and grounding performance across different event counts. Table~\ref{tab:app-comet-counting} shows that counting MAE decreases and both OBO and Pearson correlation increase for all three VLMs, indicating that the gains also extend to the accuracy of event counting.

Table~\ref{tab:app-comet-rejection} further reports Rejection-F1, the false positive rate on negative queries, and positive-query coverage. All three VLMs achieve higher Rejection-F1 and positive-query coverage after evolution. For Qwen3.5-9B and Qwen3.5-27B, these gains are accompanied by a lower false positive rate.

\begin{table}[!htb]
\centering
\AppTableCaptionStyle
\caption{\textbf{Rejection and positive-query coverage results on CoMET-Bench.} Rejection-F1 is reported on a 0--100 scale, while FPR and PosCoverage use a 0--1 scale. \textbf{Bold}: better results; \textcolor{AcademicBlue}{blue}: gains over the base skill. Evolution Gain reports absolute changes, with arrows indicating relative changes.}
\label{tab:app-comet-rejection}
\begingroup\AppTableStyle
\begin{tabularx}{\linewidth}{@{}Q{72pt}|>{\raggedright\arraybackslash}m{72pt}|*{3}{Z}}
\toprule
{\AppTableLabel\textbf{VLM}} & {\AppTableLabel\textbf{Method}} & {\bfseries \AppTableSecondary Rejection-F1 $\uparrow$} & {\bfseries \AppTableSecondary FPR $\downarrow$} & {\bfseries \AppTableSecondary PosCoverage $\uparrow$} \\ \midrule
\cellcolor{white}\multirow{4}{=}{\centering\AppTableLabel Qwen3.5-9B} & \cellcolor{gray!15}+ Base Skill & \cellcolor{gray!15}{\AppTableFont 67.51} & \cellcolor{gray!15}{\AppTableFont 0.0989} & \cellcolor{gray!15}{\AppTableFont 0.5397} \\
 & \textbf{+ Evolved Skill} & {\textbf{\AppTableFont 68.80}} & {\textbf{\AppTableFont 0.0968}} & {\textbf{\AppTableFont 0.5556}} \\
 & \textbf{Evolution Gain} & {\color{AcademicBlue}\textbf{\AppTableFont +1.29}} & {\color{AcademicBlue}\textbf{\AppTableFont $-$0.0021}} & {\color{AcademicBlue}\textbf{\AppTableFont +0.0159}} \\[-3.5pt]
 &  & {\color{AcademicBlue}\textbf{\AppTableSecondary$\uparrow$1.9\%}} & {\color{AcademicBlue}\textbf{\AppTableSecondary$\downarrow$2.1\%}} & {\color{AcademicBlue}\textbf{\AppTableSecondary$\uparrow$2.9\%}} \\
\midrule
\cellcolor{white}\multirow{4}{=}{\centering\AppTableLabel Qwen3.5-27B} & \cellcolor{gray!15}+ Base Skill & \cellcolor{gray!15}{\AppTableFont 66.13} & \cellcolor{gray!15}{\AppTableFont 0.1183} & \cellcolor{gray!15}{\AppTableFont 0.5291} \\
 & \textbf{+ Evolved Skill} & {\textbf{\AppTableFont 77.06}} & {\textbf{\AppTableFont 0.1032}} & {\textbf{\AppTableFont 0.6755}} \\
 & \textbf{Evolution Gain} & {\color{AcademicBlue}\textbf{\AppTableFont +10.93}} & {\color{AcademicBlue}\textbf{\AppTableFont $-$0.0151}} & {\color{AcademicBlue}\textbf{\AppTableFont +0.1464}} \\[-3.5pt]
 &  & {\color{AcademicBlue}\textbf{\AppTableSecondary$\uparrow$16.5\%}} & {\color{AcademicBlue}\textbf{\AppTableSecondary$\downarrow$12.8\%}} & {\color{AcademicBlue}\textbf{\AppTableSecondary$\uparrow$27.7\%}} \\
\midrule
\cellcolor{white}\multirow{4}{=}{\centering\AppTableLabel Qwen3.6-27B} & \cellcolor{gray!15}+ Base Skill & \cellcolor{gray!15}{\AppTableFont 68.53} & \cellcolor{gray!15}{\textbf{\AppTableFont 0.0989}} & \cellcolor{gray!15}{\AppTableFont 0.5529} \\
 & \textbf{+ Evolved Skill} & {\textbf{\AppTableFont 72.37}} & {\AppTableFont 0.1054} & {\textbf{\AppTableFont 0.6076}} \\
 & \textbf{Evolution Gain} & {\color{AcademicBlue}\textbf{\AppTableFont +3.84}} & {\AppTableFont +0.0065} & {\color{AcademicBlue}\textbf{\AppTableFont +0.0547}} \\[-3.5pt]
 &  & {\color{AcademicBlue}\textbf{\AppTableSecondary$\uparrow$5.6\%}} & {\AppTableSecondary$\uparrow$6.6\%} & {\color{AcademicBlue}\textbf{\AppTableSecondary$\uparrow$9.9\%}} \\
\bottomrule
\end{tabularx}
\endgroup
\end{table}

Grouped by the ground-truth event count, the results in Table~\ref{tab:app-comet-event-counts} reveal differences in multi-event grounding. Qwen3.5-9B and Qwen3.5-27B improve across all groups, while Qwen3.6-27B improves in four of the five groups. Queries with more events still exhibit lower F1@0.5, indicating that precisely detecting and localizing multiple target events in long videos remains challenging.

\begin{table}[!htb]
\centering
\AppTableCaptionStyle
\caption{\textbf{Grounding results by target-event count on CoMET-Bench.} Groups are defined by the ground-truth event count, and each column reports macro-averaged F1@0.5 over the corresponding subset of positive queries. \textbf{Bold}: better results; \textcolor{AcademicBlue}{blue}: gains over the base skill. Evolution Gain reports absolute changes, with arrows indicating relative changes.}
\label{tab:app-comet-event-counts}
\begingroup\AppTableStyle
\begin{tabularx}{\linewidth}{@{}Q{72pt}|>{\raggedright\arraybackslash}m{72pt}|*{5}{Z}}
\toprule
{\AppTableLabel\textbf{VLM}} & {\AppTableLabel\textbf{Method}} & {\bfseries \AppTableSecondary 1 event $\uparrow$} & {\bfseries \AppTableSecondary 2--3 $\uparrow$} & {\bfseries \AppTableSecondary 4--7 $\uparrow$} & {\bfseries \AppTableSecondary 8--15 $\uparrow$} & {\bfseries \AppTableSecondary $\geq$16 $\uparrow$} \\ \midrule
\cellcolor{white}\multirow{4}{=}{\centering\AppTableLabel Qwen3.5-9B} & \cellcolor{gray!15}+ Base Skill & \cellcolor{gray!15}{\AppTableFont 0.1055} & \cellcolor{gray!15}{\AppTableFont 0.0673} & \cellcolor{gray!15}{\AppTableFont 0.0460} & \cellcolor{gray!15}{\AppTableFont 0.0376} & \cellcolor{gray!15}{\AppTableFont 0.0124} \\
 & \textbf{+ Evolved Skill} & {\textbf{\AppTableFont 0.1314}} & {\textbf{\AppTableFont 0.1296}} & {\textbf{\AppTableFont 0.0574}} & {\textbf{\AppTableFont 0.0622}} & {\textbf{\AppTableFont 0.0239}} \\
 & \textbf{Evolution Gain} & {\color{AcademicBlue}\textbf{\AppTableFont +0.0259}} & {\color{AcademicBlue}\textbf{\AppTableFont +0.0623}} & {\color{AcademicBlue}\textbf{\AppTableFont +0.0114}} & {\color{AcademicBlue}\textbf{\AppTableFont +0.0246}} & {\color{AcademicBlue}\textbf{\AppTableFont +0.0115}} \\[-3.5pt]
 &  & {\color{AcademicBlue}\textbf{\AppTableSecondary$\uparrow$24.5\%}} & {\color{AcademicBlue}\textbf{\AppTableSecondary$\uparrow$92.6\%}} & {\color{AcademicBlue}\textbf{\AppTableSecondary$\uparrow$24.8\%}} & {\color{AcademicBlue}\textbf{\AppTableSecondary$\uparrow$65.4\%}} & {\color{AcademicBlue}\textbf{\AppTableSecondary$\uparrow$92.7\%}} \\
\midrule
\cellcolor{white}\multirow{4}{=}{\centering\AppTableLabel Qwen3.5-27B} & \cellcolor{gray!15}+ Base Skill & \cellcolor{gray!15}{\AppTableFont 0.2058} & \cellcolor{gray!15}{\AppTableFont 0.1846} & \cellcolor{gray!15}{\AppTableFont 0.1257} & \cellcolor{gray!15}{\AppTableFont 0.0860} & \cellcolor{gray!15}{\AppTableFont 0.0226} \\
 & \textbf{+ Evolved Skill} & {\textbf{\AppTableFont 0.2615}} & {\textbf{\AppTableFont 0.2101}} & {\textbf{\AppTableFont 0.1498}} & {\textbf{\AppTableFont 0.0939}} & {\textbf{\AppTableFont 0.0529}} \\
 & \textbf{Evolution Gain} & {\color{AcademicBlue}\textbf{\AppTableFont +0.0557}} & {\color{AcademicBlue}\textbf{\AppTableFont +0.0255}} & {\color{AcademicBlue}\textbf{\AppTableFont +0.0241}} & {\color{AcademicBlue}\textbf{\AppTableFont +0.0079}} & {\color{AcademicBlue}\textbf{\AppTableFont +0.0303}} \\[-3.5pt]
 &  & {\color{AcademicBlue}\textbf{\AppTableSecondary$\uparrow$27.1\%}} & {\color{AcademicBlue}\textbf{\AppTableSecondary$\uparrow$13.8\%}} & {\color{AcademicBlue}\textbf{\AppTableSecondary$\uparrow$19.2\%}} & {\color{AcademicBlue}\textbf{\AppTableSecondary$\uparrow$9.2\%}} & {\color{AcademicBlue}\textbf{\AppTableSecondary$\uparrow$134.1\%}} \\
\midrule
\cellcolor{white}\multirow{4}{=}{\centering\AppTableLabel Qwen3.6-27B} & \cellcolor{gray!15}+ Base Skill & \cellcolor{gray!15}{\AppTableFont 0.2085} & \cellcolor{gray!15}{\AppTableFont 0.1597} & \cellcolor{gray!15}{\AppTableFont 0.1342} & \cellcolor{gray!15}{\textbf{\AppTableFont 0.0990}} & \cellcolor{gray!15}{\AppTableFont 0.0322} \\
 & \textbf{+ Evolved Skill} & {\textbf{\AppTableFont 0.2578}} & {\textbf{\AppTableFont 0.1800}} & {\textbf{\AppTableFont 0.1378}} & {\AppTableFont 0.0901} & {\textbf{\AppTableFont 0.0377}} \\
 & \textbf{Evolution Gain} & {\color{AcademicBlue}\textbf{\AppTableFont +0.0493}} & {\color{AcademicBlue}\textbf{\AppTableFont +0.0203}} & {\color{AcademicBlue}\textbf{\AppTableFont +0.0036}} & {\AppTableFont $-$0.0089} & {\color{AcademicBlue}\textbf{\AppTableFont +0.0055}} \\[-3.5pt]
 &  & {\color{AcademicBlue}\textbf{\AppTableSecondary$\uparrow$23.6\%}} & {\color{AcademicBlue}\textbf{\AppTableSecondary$\uparrow$12.7\%}} & {\color{AcademicBlue}\textbf{\AppTableSecondary$\uparrow$2.7\%}} & {\AppTableSecondary$\downarrow$9.0\%} & {\color{AcademicBlue}\textbf{\AppTableSecondary$\uparrow$17.1\%}} \\
\bottomrule
\end{tabularx}
\endgroup
\end{table}

\begin{figure}[p]
\centering
\includegraphics[page=1,width=\linewidth]{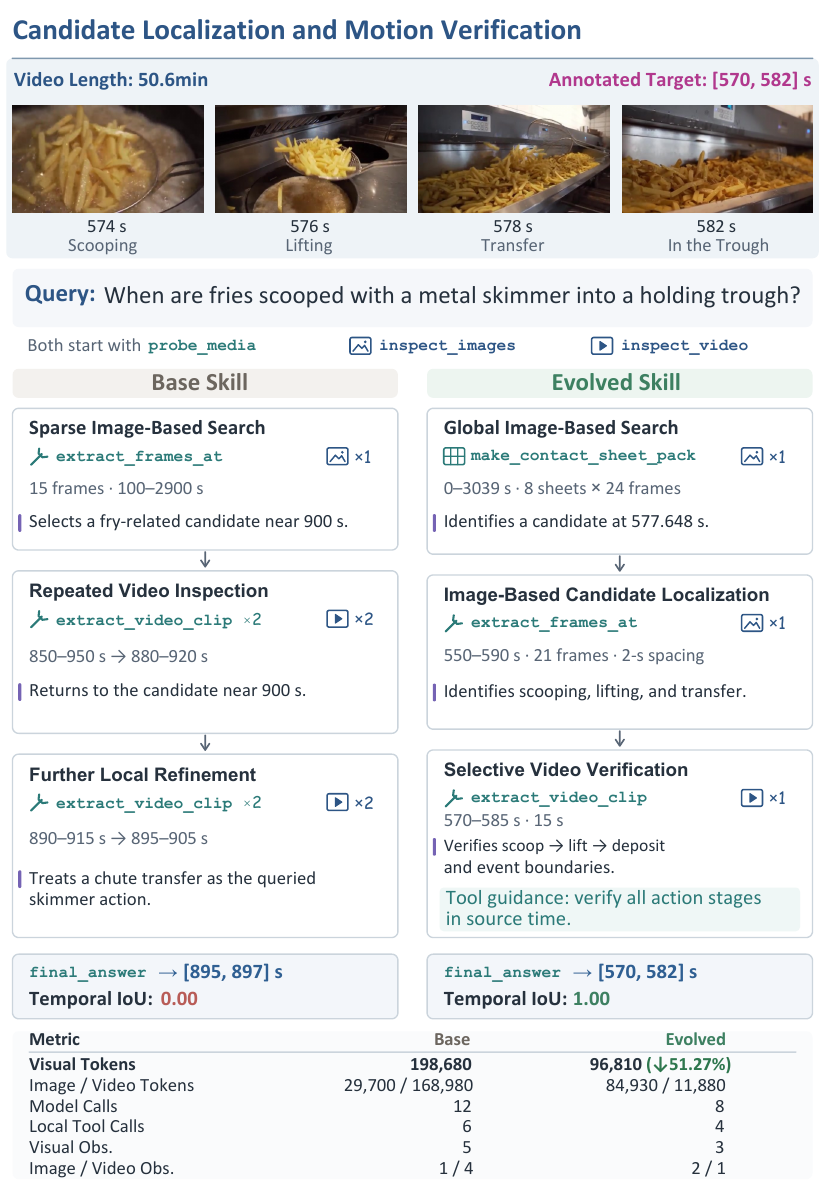}
\caption{\textbf{Candidate localization and motion verification.} The base skill repeatedly inspects an incorrect candidate, whereas the evolved skill combines global contact-sheet search with local frame inspection and a single short video to verify the complete action, precisely localizing the target interval while reducing visual token cost by \textbf{51.27\%}.}
\label{fig:case-motion}
\end{figure}

\begin{figure}[p]
\centering
\includegraphics[page=2,width=\linewidth]{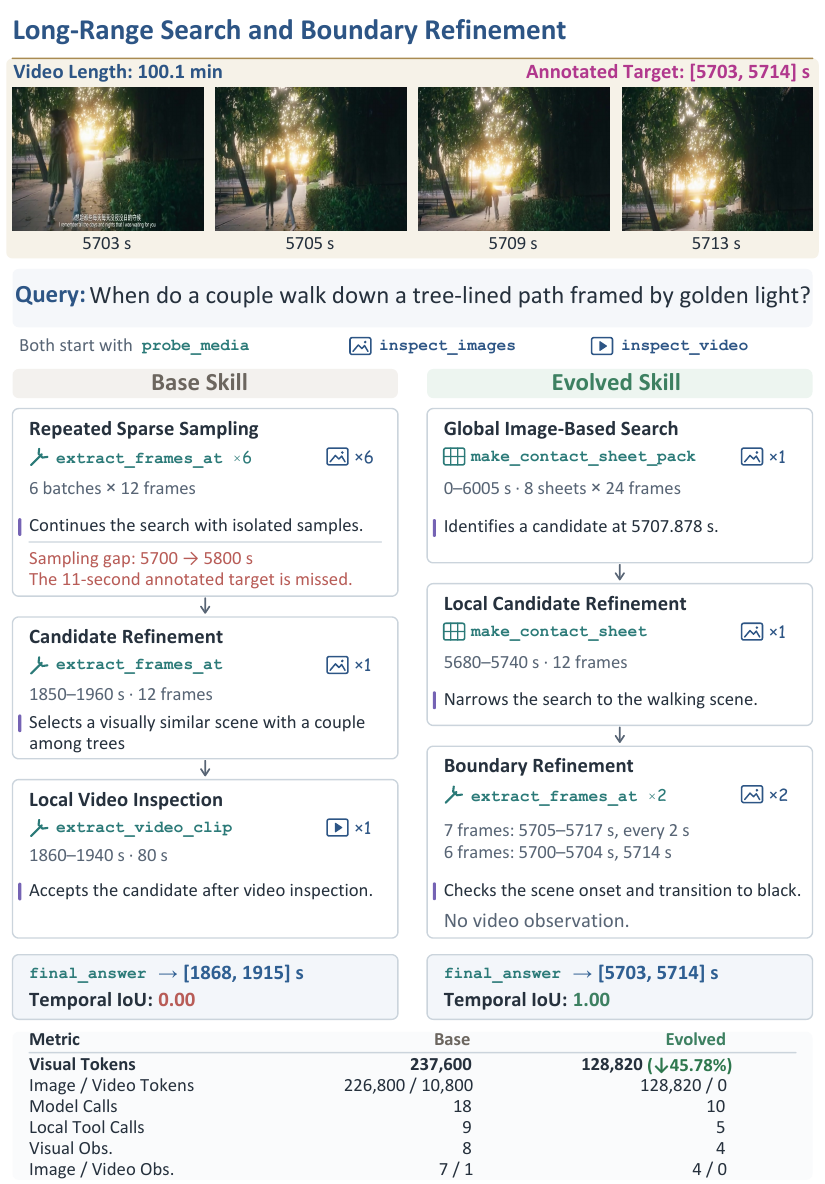}
\caption{\textbf{Long-range search and boundary refinement.} Sparse sampling misses the brief target, leading the base skill to mistakenly focus on a visually similar scene. The evolved skill instead combines global and local contact sheets with boundary frames to precisely localize the target interval, reducing visual token cost by \textbf{45.78\%}.}
\label{fig:case-search}
\end{figure}

\raggedbottom
\begin{samepage}
\section{Case Study}
\label{app:cases}

We analyze the execution trajectories of the base and evolved skills on two ExtremeWhenBench~\citep{seo2026extremewhenbench} queries using the same Qwen3.5-27B. Figures~\ref{fig:case-motion} and~\ref{fig:case-search} illustrate how policy--tool coevolution changes evidence acquisition for localizing an action and a scene, respectively. Both skills remain fixed during evaluation, and visual token costs are computed following the definitions in Appendix~\ref{app:metrics}.

\end{samepage}

In Figure~\ref{fig:case-motion}, the query asks when fries are scooped with a metal skimmer into a holding trough in a 50.6-minute video. The base skill selects a candidate near 900\,s from 15 sparse frames and inspects four overlapping clips in this region. Although the initial clips do not confirm the queried action, the model continues to inspect the same region and ultimately mistakes a chute transfer for the requested action. In contrast, the evolved skill uses timestamped contact sheets for global comparison, followed by dense frame inspection over 550--590\,s and a single 15-second clip for verification. The prediction matches the ground-truth interval $[570,582]$\,s, raising IoU from 0 to 1. Meanwhile, video tokens drop from 168.98k to 11.88k, leading to a \textbf{51.27\%} net reduction in visual token cost. This trajectory illustrates how better candidate localization allows video observation to focus on verifying query-specific motion, reducing repeated inspection of an incorrect region.

Figure~\ref{fig:case-search} concerns an 11-second scene of a couple walking along a tree-lined path in golden light within a 100.1-minute video. The base skill performs six batches of sparse sampling across the long timeline but misses the brief target scene. Subsequent frame and video inspection focuses on a visually similar candidate, yielding the incorrect interval $[1868,1915]$\,s. The evolved skill instead packs 192 frames into eight contact sheets, allowing a broad search within one image observation. After identifying a candidate near 5708\,s, it uses a local contact sheet over 5680--5740\,s and targeted boundary frames to determine when the scene begins and ends. These image-based observations localize the target to $[5703,5714]$\,s without video observation, increasing IoU from 0 to 1 while reducing visual token cost by \textbf{45.78\%}. This case highlights the importance of discovering a brief target during global search before investing in local boundary refinement.

Together, these two cases illustrate the connection between tool capabilities and observation orchestration: the evolved tools provide compact temporal coverage and explicit temporal correspondence, while the policies organize subsequent observations around the evidence still needed. In the base trajectories, the additional local evidence gathered remains confined to an incorrect candidate. The evolved trajectories instead improve candidate localization before resolving the remaining temporal uncertainty, using video to verify action progression in one case and images to determine scene boundaries in the other. This contrast illustrates how coordinated image--video observation adapts to the evidence needs of different queries. By coupling the form in which evidence is presented with decisions about where to search and what to verify, the evolved skill enables the VLM to achieve more precise grounding at a lower visual token cost.


\clearpage

\section{Limitations and Future Work}
\label{app:limitations}

At present, our policy--tool coevolution is conducted on a limited set of temporal grounding queries from a single data source with labeled feedback. Although the evolved skill improves performance across ultra-long video temporal grounding and long-video QA tasks, broader data sources, larger evolution scales, and evolution without labeled feedback remain to be further explored. All visual observations in the current framework are performed directly by the VLM acting as the main agent. A meaningful next step is to extend coevolution to determine when the main agent should observe directly or delegate observation to subagents, and how they can coordinate efficiently. Since we currently focus only on the visual modality, extending the framework to more general omni-modal settings also merits investigation.

\raggedbottom
\section{Prompts}
\label{app:prompts}

\subsection{Skill Prompts}
\label{app:skill-prompts}

\begin{promptbox}[unbreakable]{System Prompt}
\noindent You are a video-task solving agent.\par
\smallskip
\noindent Active evolvable skill:\par
\noindent \textless{}SKILL.md, tool-selection.md, tool-notes.md, policy-notes.md\textgreater{}\par
\smallskip
\noindent Current task adapter:\par
\noindent \textless{}optional task-specific adapter\textgreater{}\par
\end{promptbox}
\begin{promptbox}{Base SKILL.md}
\noindent This skill supports ultra-long video temporal grounding with image and video observations.\par
\promptheading{Evidence acquisition}
\begin{promptsteps}
  \item Use local tools to read video metadata and prepare visual evidence. \texttt{probe\_media} reads metadata, \texttt{extract\_video\_clip} prepares video clips, and \texttt{extract\_frames\_at} prepares timestamped frames.
  \item Choose the observation tool for the current evidence need:
  \begin{promptitems}
    \item \texttt{inspect\_video} for native video observation of motion, event order, state changes, fine details, or temporal boundaries.
    \item \texttt{inspect\_images} for native image observation of static detail, readable text, object identity, or boundary frames.
  \end{promptitems}
  \item Use source-video seconds in the final temporal answer.
\end{promptsteps}
\promptheading{Fixed constraints}
\begin{promptitems}
  \item Local tools prepare visual evidence and metadata. The VLM interprets the evidence and predicts temporal intervals.
  \item Do not call external models or APIs from local tools.
  \item Do not use audio, speech recognition, transcripts, hidden labels, filenames, or ground truth as evidence.
  \item Submit the final prediction with \texttt{final\_answer} only after collecting sufficient visual evidence.
\end{promptitems}
\smallskip
\noindent No observation orchestration policy has been learned through evolution yet.\par
\end{promptbox}

\subsection{Evaluation Task Prompts}
\label{app:evaluation-prompts}
\label{app:grounding-prompts}
\label{app:qa-prompt}

\begin{promptbox}[unbreakable]{VUE-LVTR Prompt}
\noindent Task: \textless{}text query\textgreater{}\par
\smallskip
\noindent Required answer format: a JSON array containing all matching [start\_seconds, end\_seconds] intervals.\par
\smallskip
\noindent Declared video duration: \textless{}duration\textgreater{} seconds.\par
\smallskip
\noindent Use source-video seconds for temporal answers.\par
\end{promptbox}
\begin{promptbox}[unbreakable]{ExtremeWhenBench Prompt}
\noindent You are watching a video of \textless{}duration\textgreater{} seconds. Please find the moment described by the following question, determining its starting and ending times.\par
\smallskip
\noindent Question: \textless{}question\textgreater{}\par
\smallskip
\noindent Required answer format: a single [start\_seconds, end\_seconds] interval in source-video seconds.\par
\end{promptbox}
\begin{promptbox}[unbreakable]{CoMET-Bench Prompt}
\noindent Watch the provided video and answer the following question:\par
\smallskip
\noindent \textless{}query\textgreater{}\par
\smallskip
\noindent Return ONLY a JSON array of temporal intervals where the event occurs, each formatted as [start\_seconds, end\_seconds]. Return [] if the event does not occur. No explanation.\par
\end{promptbox}

\begin{promptbox}[unbreakable]{Long-Video QA Prompt}
\noindent The current task is multiple-choice video question answering. Reuse the active skill to locate and verify the relevant visual evidence. Any temporal-interval final-output convention in the skill does not apply to this task. Finish by returning exactly one option letter: A, B, C, or D.\par
\smallskip
\noindent \textless{}question\textgreater{}\par
\noindent \textless{}answer options\textgreater{}\par
\end{promptbox}

\clearpage
\subsection{Codex Updater Prompt}
\label{app:updater-prompt}

\begin{promptbox}[unbreakable]{Codex Updater Prompt}
\noindent You are optimizing a compact reusable skill for ultra-long video temporal grounding.\par
\promptheading{Inputs}
\begin{promptitems}
  \item Update context with trajectory summaries, task feedback, and evaluation metrics: \textless{}update context path\textgreater{}
  \item Execution trajectories in the current batch:
  \begin{promptitems}
    \item Episode \textless{}episode index\textgreater{}: \textless{}trajectory path\textgreater{}
    \item[] \ldots
  \end{promptitems}
  \item Batch feedback and aggregate metrics: \textless{}batch feedback path\textgreater{}
  \item Previous skill (read-only): \textless{}previous skill directory\textgreater{}
  \item Candidate skill (a copy of the previous skill to update in place): \textless{}candidate skill directory\textgreater{}
\end{promptitems}
\promptheading{Evolution rules}
\begin{promptitems}
  \item Edit only the candidate skill. Keep the previous skill, trajectories, and feedback unchanged. You may edit \path{SKILL.md}, \path{tools/}, and the three reference documents listed below. All editable paths are relative to the candidate skill directory.
  \item First analyze all execution trajectories in the batch: execution steps, tool choices, visual observations, model--tool interactions, common failures, and successful patterns.
  \item Use the update context for evaluation metrics and final outcomes. Do not copy sample identifiers, video or file names, question text, scene facts, predictions, ground-truth intervals, dataset-specific labels or groupings, dataset-specific patterns, or content-type examples into the skill. Express reusable lessons as task-general evidence workflows, tool contracts, validation checks, or guidance for model--tool interactions.
  \item Keep local tools deterministic and local. They prepare visual evidence and metadata for the VLM, which interprets the evidence and predicts temporal intervals. Do not call external models or APIs, or use audio, speech recognition, or transcripts.
  \item Record reusable lessons in the appropriate skill files:
  \begin{promptitems}
    \item \path{references/policy-notes.md}: task planning, evidence acquisition, and reasoning strategies.
    \item \path{references/tool-notes.md}: tool contracts, failure modes, and usage details.
    \item \path{references/tool-selection.md}: selection of local tools and observation modes.
  \end{promptitems}
  Consider the trade-offs between native image and video observation, including static detail, readable text, object identity, and boundary-frame verification with \texttt{inspect\_images}.
  \item Ground updates in the listed execution trajectories: tool choices, visual readability, temporal coverage, interaction failures, token usage, and repeated patterns where a better reusable tool or tool-selection rule would help.
  \item Analyze token usage as part of skill quality, not just task score. Prefer updates that improve accuracy while keeping token use stable or lower. Add broad video inspection, longer clips, or extra observation calls only when trajectories support a reusable accuracy benefit that justifies the cost.
  \item Use code synthesis for reusable improvements. Add or improve local media-preparation tools for clipping, frame sampling, resizing, timestamp labeling, compact visual summaries, motion- or scene-aware media preparation, and validation. Merge or delete weak or redundant tools. Store each tool's code and interface in \path{tools/<tool_name>/tool.py} and \path{tools/<tool_name>/tool.json}.
  \item Consolidate or replace weak guidance. Leave the candidate unchanged if there is no reusable improvement.
\end{promptitems}
\end{promptbox}

\end{document}